\documentclass[pdflatex,sn-basic]{sn-jnl}

\usepackage{amssymb}
\usepackage{graphicx}
\usepackage{url}
\usepackage[labelsep=endash, figurename=Figure, tablename=Table, textfont=it]{caption}
\usepackage{subcaption}
\usepackage{float}  
\usepackage{amsmath}
\usepackage{hyperref, wasysym}
\usepackage{multicol}
\usepackage{xcolor}
\usepackage{booktabs}
\usepackage{rotating}
\usepackage{multirow}
\usepackage{algorithm}%
\usepackage{algorithmicx}%
\usepackage{algpseudocode}%
\usepackage{fancyhdr}
\usepackage{titlesec}
\usepackage{amsthm}
\newtheorem{definition}{Definition}
\usepackage{booktabs}

\usepackage{listings}

\theoremstyle{thmstyleone}%
\theoremstyle{thmstyletwo}%

\theoremstyle{thmstylethree}%

\begin{document}

    
    
    

    

\title[Article Title]{Robustness of Reinforcement Learning-Based Traffic Signal Control under Incidents: A Comparative Study}


\author*[1]{\fnm{Dang Viet Anh} \sur{Nguyen}}\email{andng@dtu.dk}

\author[1]{\fnm{Carlos Lima} \sur{Azevedo}}\email{climaz@dtu.dk}

\author[2]{\fnm{Tomer} \sur{Toledo}}\email{toledo@technion.ac.il}

\author[1]{\fnm{Filipe} \sur{Rodrigues}}\email{rodr@dtu.dk}

\affil*[1]{\orgdiv{Department of Technology, Management, and Economics}, \orgname{Technical University of Denmark (DTU)}, \orgaddress{\postcode{2800} \city{Kongens Lyngby}, \country{Denmark}}}

\affil[2]{\orgdiv{Faculty of Civil and Environmental Engineering}, \orgname{Technion – Israel Institute of Technology}, \orgaddress{\city{Haifa} \postcode{32000}, \country{Israel}}}


\abstract{Reinforcement learning-based traffic signal control (RL-TSC) has emerged as a promising approach for improving urban mobility. However, its robustness under real-world disruptions such as traffic incidents remains largely underexplored. In this study, we introduce T-REX, an open-source, SUMO-based simulation framework for training and evaluating RL-TSC methods under dynamic, incident scenarios. T-REX models realistic network-level performance considering drivers' probabilistic rerouting, speed adaptation, and contextual lane-changing, enabling the simulation of congestion propagation under incidents. To assess robustness, we propose a suite of metrics that extend beyond conventional traffic efficiency measures. Through extensive experiments across synthetic and real-world networks, we showcase T-REX for the evaluation of several state-of-the-art RL-TSC methods under multiple real-world deployment paradigms. Our findings show that while independent value-based and decentralized pressure-based methods offer fast convergence and generalization in stable traffic conditions and homogeneous networks, their performance degrades sharply under incident-driven distribution shifts. In contrast, hierarchical coordination methods tend to offer more stable and adaptable performance in large-scale, irregular networks, benefiting from their structured decision-making architecture. However, this comes with the trade-off of slower convergence and higher training complexity. These findings highlight the need for robustness-aware design and evaluation in RL-TSC research. T-REX contributes to this effort by providing an open, standardized and reproducible platform for benchmarking RL methods under dynamic and disruptive traffic scenarios.}


\maketitle



\section{Introduction}\label{sec1}

Traffic congestion continues to be a global challenge for urban mobility, causing economic losses and reduced quality of life. In 2024, Istanbul drivers faced the highest global delays, losing an average of 105 hours in traffic—a 15\% increase from the previous year. In New York City and Chicago, motorists lost 102 hours, amounting to over \$1,800 in lost productivity per driver. In total, U.S. drivers lost more than four billion hours and \$74 billion annually to congestion \citep{INRIX2024}. Effective traffic signal control is crucial for improving traffic flow, reducing delays, and enhancing overall mobility.

Traditional traffic signal control methods—fixed-time, actuated, and adaptive systems such as SCOOT \citep{robertson1991optimizing}, SCATS \citep{sims1980sydney}, and TUC \citep{diakaki2002multivariable}—have long been deployed in practice. While fixed-time control is simple, it lacks responsiveness to real-time fluctuations. Actuated control offers flexibility through local sensor input, but its limited coordination across intersections hinders overall network efficiency. Adaptive systems attempt real-time coordination but often struggle with large-scale disruptions and rapidly evolving traffic conditions \citep{agarwal2024dynamic}.

Recent advancements in artificial intelligence have positioned reinforcement learning (RL) as a powerful alternative for traffic signal control (RL-TSC). RL agents optimize signal policies by interacting with traffic environments and learning long-term strategies that enhance network-wide performance \citep{noaeen2022reinforcement}. Unlike traditional methods, RL-TSC is adaptive to real-time dynamics and capable of discovering complex temporal patterns, making it well-suited for managing urban congestion and emissions. The growing availability of traffic data and computing infrastructure supports RL-TSC as a promising, scalable solution.

However, despite its theoretical strengths, RL-TSC remains largely undeployed in real-world urban networks \citep{chen2022real, han2023leveraging}. Most studies are constrained to simulated environments or historical datasets that do not capture real-time uncertainties. These simulation setups frequently assume stable traffic conditions, failing to reflect abnormal scenarios such as incidents, sensor errors, or dynamic rerouting behaviors \citep{haydari2020deep}. This disconnect between controlled experiments and real-world complexities highlights the need for rigorous robustness evaluation in RL-based traffic control.

In computer science, robustness refers to a system's ability to perform reliably under atypical or adverse conditions \citep{pullum2022review}. In control theory, it implies insensitivity to parameter variations in a closed-loop system \citep{aastrom2021feedback}. In reinforcement learning, robustness entails coping with environmental uncertainty while using all available information \citep{yamagata2024safe}. In urban traffic systems, uncertainties arise from fluctuating demand, sensor errors, accidents, weather, and other unpredictable disruptions. These changes can significantly alter traffic dynamics, posing major challenges for RL-TSC systems, which often fail to generalize outside their training distribution.

We define:

\begin{definition}
\textbf{Robustness in Reinforcement Learning-Based Traffic Signal Control (RL-TSC)} is the system’s ability to maintain stable and efficient performance under dynamic, uncertain, and adversarial conditions in the traffic network and infrastructure. A robust RL-TSC system should adapt to and generalize across diverse and unpredictable scenarios, including traffic demand fluctuations, incidents, sensor failures, and environmental disruptions, while ensuring safety, efficiency, and stability in signal control decisions.
\end{definition}

A growing body of research has investigated how to enhance robustness in RL-TSC. RL methods rely on accurate traffic state information, and sensor failures—due to hardware faults, data loss, or detection noise—can disrupt policy performance. The more complex the state representation, the more sensitive the RL controller becomes to such failures. Several studies tackle this issue directly. \cite{rodrigues2019towards} simulate sensor failures and use state aggregation, dropout-based training, and historical patterns to improve the resilience of traffic signal control at a single intersection. \cite{shi2023improving} introduce RGLight, which leverages graph neural networks (GNNs), distributional RL, and policy ensembles to maintain stability under incomplete or noisy inputs. \cite{jiang2024blindlight} propose a dual-policy system that switches between a standard RL controller and a cyclic fallback policy depending on observation quality. Other works adopt transfer learning and generalization strategies to improve adaptation to degraded or uncertain sensor inputs \citep{wu2020multi, xu2022robustness, tan2020robust, zhang2020using}.

Beyond sensing issues, demand fluctuations introduce additional stochasticity. Daily patterns, events, and unexpected congestion can cause non-stationary traffic distributions. \cite{rodrigues2019towards} demonstrate RL-TSC resilience to sudden demand surges at a single intersection, provided that the model has been exposed to such scenarios during training. \cite{shi2023improving} extend this to larger networks (e.g., Luxembourg and Manhattan), confirming improved performance under traffic volatility compared to baseline RL-TSC approaches.

More disruptive still are traffic incidents, which can block lanes or intersections, triggering sharp shifts in traffic flow. \cite{rodrigues2019towards} model incidents by halting vehicles or reducing lane availability. Their findings show that RL policies trained with incident exposure are more resilient than conventional baselines. \cite{zeinaly2023resilient} simulate vehicle stoppages at varying distances and confirm that RL agents can preserve learning stability. \cite{aslani2018traffic} use the AIMSUN simulator to model lane blockages in Tehran and find that actor-critic methods—an advanced form of RL—perform better than traditional approaches like Q-learning and SARSA in handling such disruptions. However, most of these incident models remain localized—affecting only one intersection or link—and ignore the network-wide spillover effects typical of real-world incidents. Disruptions often propagate due to rerouting, queue spillback, and congestion shockwaves, creating a systemic performance collapse across multiple intersections. Current RL-TSC research rarely addresses this full-network challenge. Moreover, while sensor failure and demand variations can be easily injected as noise or data masking, realistic incident modeling introduces distributional shifts in both traffic patterns and driver behavior—posing a unique and underexplored challenge.

To address this gap, we introduce \textbf{T-REX} (\textbf{T}raffic signal control framework for \textbf{R}obustness \textbf{E}valuation under e\textbf{X}treme scenarios)—a benchmarking framework for evaluating RL-TSC methods under disruptive traffic events. T-REX is designed to simulate network-wide disruptions by incorporating state-of-the-art traffic models that capture dynamic rerouting, speed variations, and lane-changing behaviors in response to incidents. By providing a reproducible and open-source platform, T-REX facilitates the comprehensive evaluation of the learning stability, robustness, and transferability of RL methods under various traffic uncertainties. While previous RL studies often rely on diverse evaluation protocols and metrics, this lack of standardization has led to inconsistent results and limited reproducibility \citep{jordan2020evaluating}. T-REX addresses this critical limitation by offering a principled and standardized framework for the reliable assessment of RL-TSC methods.

Our key contributions can be summarized as follows:

\begin{itemize}
\item We introduce T-REX, a open-source simulation framework designed to evaluate RL-TSC methods under traffic incident scenarios. The framework incorporates models for rerouting, speed adjustment, and contextual lane-change behavior, facilitating systematic, network-level robustness evaluation.

\item For benchmarking and reproducibility, we propose a set of evaluation metrics tailored to RL-TSC, which extend beyond conventional traffic performance indicators to capture various dimensions of control robustness.

\item To showcase the above, we conduct a comparative analysis of the robustness of state-of-the-art RL-TSC methods. We find that no single method consistently dominates across all robustness dimensions; performance varies with network topology, and exposure to incidents during training does not improve test-time performance under incidents without explicit incident-awareness. Transferability remains a key challenge, underscoring the need for methods designed to handle dynamic, real-world conditions.

\end{itemize}


The remainder of this paper is organized as follows. Section \ref{sec3:method} describes the proposed T-REX framework, detailing its architecture and key modules designed to simulate vehicle behavior under incident conditions. Section \ref{sec4:experiment} outlines the traffic networks, RL-TSC methods evaluated, parameter configurations, and the robustness metrics used. Finally, Section \ref{sec5:results} presents the experimental setup and results of our comparative analysis, focusing on learning performance, testing performance and generalization, transferability and adaptation of RL-TSC methods.

\section{Methodology}
\label{sec3:method}

\subsection{Problem definition}

Traffic signal control in urban networks involves dynamically selecting signal phases at intersections to minimize vehicle delays and congestion. The control decisions rely on data collected from traffic sensors, which provide partial and local observations of the current traffic conditions. The problem becomes more complex in the presence of incidents (e.g., lane blockages), which disrupt normal traffic patterns and introduce uncertainty in traffic flow.

We model traffic signal control under incident scenarios as a Decentralized Partially Observable Markov Decision Process (Dec-POMDP), defined by the tuple $\mathcal{M} = \langle \mathcal{S}, \mathcal{A}, P, R, \Omega, O, \gamma, \mathcal{N} \rangle$, where $\mathcal{N}$ is the set of signalized intersections, each controlled by a decentralized agent. At each time step $t$, the global traffic state is $s_t \in \mathcal{S}$, and each agent $i$ receives a local observation $o^i_t = O(s_t)$ based on its intersection’s traffic conditions and current phase. Each agent selects an action $a^i_t \in \mathcal{A}^i$—a predefined signal phase—applied for a fixed duration.

Traffic incidents are introduced at randomly selected edges $e^i \in E$, blocking lanes $l \subseteq L_i$ approaching intersection $i$. The incident duration $d$ and start time $t_{\text{start}}$ are drawn from predefined distributions $\mathcal{D}$ and $\mathcal{U}$, respectively. The transition function $P(s_{t+1} | s_t, a_t)$ models the evolution of the traffic network, capturing both regular flow and incident-related disruptions, such as rerouting, speed reduction, and lane-changing behavior. These dynamics are simulated using the T-REX framework, providing a realistic and data-driven representation of incident impacts.

Each agent aims to minimize congestion and delay through the reward function $R$. The goal is to learn an optimal joint policy:
\begin{align}
    \pi(\boldsymbol{a}_t | \boldsymbol{s}_t) = \prod_{i \in \mathcal{N}} \pi^i_\theta(a^i_t | o^i_t),
\end{align}
that maximizes the expected discounted return:
\begin{align}
    J(\pi) = \mathbb{E}_{\tau \sim \pi} \left[ \sum_{t=0}^\infty \sum_{i \in \mathcal{N}} \gamma^t r^i_t \right],
\end{align}
where $\theta_i$ denotes the parameters of agent $i$’s policy, and $\tau$ is the trajectory over states and actions. Since learning a centralized policy over the joint state-action space suffers from the curse of dimensionality, we instead learn a decentralized policy $\pi = \{\pi_i\}_{i \in \mathcal{N}}$, where each agent independently maximizes its own long-term cumulative reward.

\subsection{Overall Framework}

\begin{figure}
	\centering
	\includegraphics[width=\textwidth]{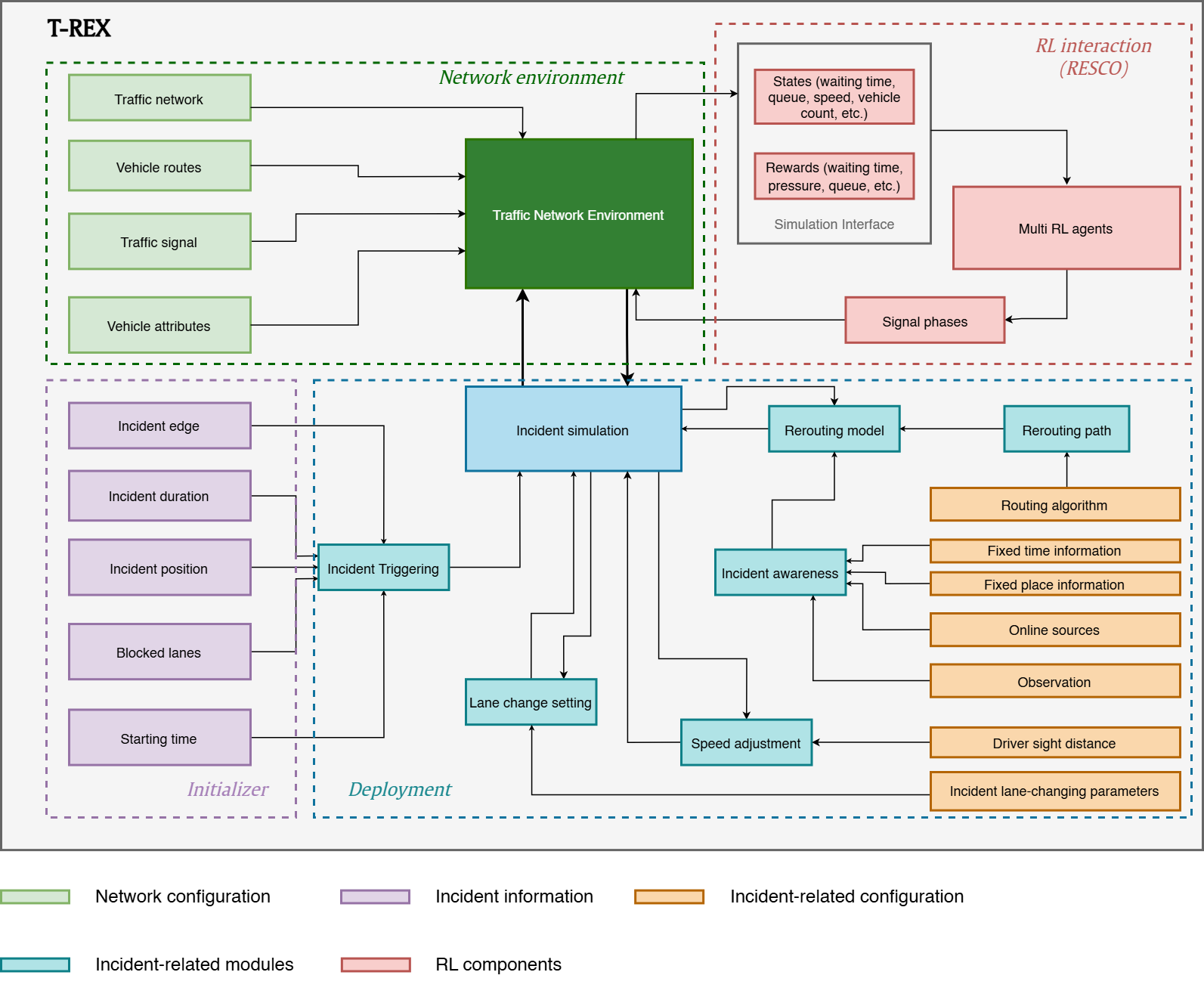}
	\caption{Schematic overview of the T-REX framework}
    \label{sec3:t-rex}
\end{figure}

T-REX is a simulation-based framework developed to model and analyze traffic signal control at the network level under incident scenarios. It is designed primarily as an offline tool to support the training and evaluation of control policies—particularly reinforcement learning-based methods—by providing realistic traffic dynamics that include disruptions such as lane blockages, rerouting behavior, and speed adaptation. Beyond RL, T-REX can also be used to evaluate alternative traffic management strategies, sensor configurations, and control logics in a controlled, reproducible setting. In this study, we use T-REX to systematically train and evaluate RL-TSC methods under incident conditions. A schematic overview of the framework is provided in Figure~\ref{sec3:t-rex}.

Built on the open-source SUMO simulator \citep{lopez2018microscopic}, T-REX comprises four main modules: \textit{Network Environment}, \textit{Initializer}, \textit{Deployment}, and \textit{RL Interaction}. The \textit{Network Environment} module forms the SUMO-based simulation core, initializing the scenario with network layout, vehicle routes, signal plans, and vehicle attributes. The \textit{Initializer} defines incident parameters—either user-defined or randomly generated—including location, duration, blocked lanes, and start time. The \textit{Deployment} module injects incidents into the simulation and adjusts vehicle behavior accordingly, such as rerouting, lane changes, and speed adaptation. The \textit{RL Interaction} module enables real-time communication between agents and the simulation, transmitting traffic states and rewards while receiving control actions (signal phases).

T-REX uses the TraCI interface for real-time simulation control, enabling dynamic observation, incident triggering, and manipulation of vehicle behaviors during execution. Further implementation details are described in the following sections.

\subsection{Incident Definition and Deployment}

Traffic incidents—such as collisions, roadwork, or severe weather—are unexpected events that disrupt normal traffic flow by blocking part of a road for a certain duration. Although SUMO is widely used for traffic simulation, it lacks built-in support for modeling such incidents. Existing workarounds typically involve stopping vehicles, altering traffic light phases, or reducing road capacity or speed limits \citep{li2022python, smith2014sumo}. However, these approaches are limited: capacity or speed reductions lack realism, traffic light manipulation cannot simulate mid-edge disruptions, and manually defining affected areas becomes impractical for large or randomized networks. Furthermore, stopping vehicles using predefined routes fails to support stochastic or on-the-fly incident generation, making it unsuitable for RL training. Fixed-duration incidents also fail to capture the variability of real-world clearance times.

To address these limitations, we propose an online incident deployment method using TraCI. Incident parameters are defined in the \textit{Initializer} module, supporting both user-defined and random generation. Each incident is specified by its edge $e^d \in E$, position $p$, blocked lanes $l \subseteq L_i$, start time $t_{\text{start}}$, and duration $d$. For randomized deployment, the number of blocked lanes is sampled from a uniform distribution $l \sim \mathcal{U}(1, l_n)$, where $l_n$ is the number of lanes on edge $e^d$, and the position $p$ is sampled uniformly within the range (10, $x_e$), where $x_e$ is the edge length, and the 10-meter buffer at each end prevents bugs in SUMO. The duration $d$ is sampled from a distribution $\mathcal{D}$, and $t_{\text{start}}$ is sampled uniformly from $(t_{\text{warmup}}, t_{\text{end}} - 1200)$ ensuring incidents start after warm-up and have sufficient time to affect traffic before the simulation ends. Multiple incidents can be configured to occur simultaneously.

Incident activation is handled by the \textit{Deployment} module. We first explored collision-based triggering using TraCI, where vehicles are deliberately slowed to simulate collisions at position $p$. While this method realistically simulates accidents, it introduces timing variability due to vehicle availability ($t_{\text{start}} + \Delta t$), thus affecting evaluation metrics, and it is limited to collision scenarios.

To overcome these issues, we introduce a more flexible method: at $t_{\text{start}}$, special ``IC'' vehicles are inserted at position $p$ on edge $e^d$ to block the designated lanes. If a vehicle is already present, it is teleported to maintain consistency. This approach offers precise control over timing and location, ensures immediate blockage, simplifies implementation, and better supports realistic, randomized incident scenarios during RL training.

\subsection{Simulating vehicle behavior under incident}

\subsubsection{Rerouting model}
Existing studies on RL-TSC under incident scenarios generally do not incorporate vehicle rerouting into their simulations. In previous work, \cite{smith2014sumo} proposed a rerouting mechanism while developing a simulation for incidents in SUMO. However, their rerouting logic is relatively simplistic—it reroutes any vehicle upstream of the incident edge if that edge appears in its planned route. This approach does not accurately reflect real-world rerouting behavior. In reality, rerouting is a far more complex phenomenon. Drivers’ awareness of an incident is influenced by various information sources, such as radio, navigation systems, road signage, and direct observation. The timing and decision to reroute also vary greatly among individuals. Neglecting this complexity in simulation limits the realism of traffic dynamics under incident scenarios, which can lead to misleading conclusions about the robustness and effectiveness of RL-TSC methods.

To more realistically simulate rerouting behavior in our framework, we adopt the Information Comply Model (ICM) \citep{kucharski2019simulation} and implement it in SUMO using TraCI. The ICM is built on three key assumptions. First, drivers only consider rerouting once they become aware of the incident, which depends on both direct observation and external information sources. Second, drivers choose to reroute to avoid the negative consequences of the event and to benefit from expected lower travel costs. Third, when rerouting, drivers aim for an optimal path based on their estimates of actual and future travel costs, which may be influenced by other rerouting drivers, past experiences, or predictive traffic information.

The ICM models driver behavior using two components: an awareness model, which determines when and how drivers become aware of an incident, and a rerouting model, which governs the decision-making process for selecting an alternative route.

The awareness model aims to represent how drivers become informed about unexpected events based on various information sources. These sources include fixed-time information (FTI), such as radio broadcasts and periodic TMC updates; fixed-place information (FPI), such as roadside Variable Message Signs (VMS); online sources (OS), such as social media, websites, or mobile applications; and direct observation (OB) of abnormal traffic conditions. 

Once drivers become aware of an unexpected event, the rerouting model is used to estimate the probability that they will choose to deviate from their original route. This probability is derived using a binomial logit model, where the utility of rerouting is determined by the expected gain and avoided loss associated with the decision. 

Further details on the ICM components are provided in Appendix \ref{secA2}.

\subsubsection{Speed adaptation model}

When a vehicle enters an edge affected by an incident, drivers typically reduce their speed due to safety concerns, environmental factors, and behavioral responses such as rubbernecking \citep{cao2021quantification}. Visibility, road geometry, weather, and traffic density influence the extent of this slowdown. In our simulation, we model this behavior through speed adaptation based on the driver's visibility of the incident ahead.

We propose a novel method based on the concept of Stopping Sight Distance (SSD) from road design guidelines \citep{aashto2018green} to determine the point at which a driver initiates deceleration in our simulation. In this context, SSD refers to the total distance a vehicle needs to come to a complete stop, accounting for both the perception-reaction distance and the braking distance. The perception-reaction distance (m) is computed as:
\[
d_{\text{PRT}} = 0.278 \times V \times t,
\]
where \( V \) is the vehicle speed in km/h and \( t = 2.5 \) seconds, the standard reaction time recommended by AASHTO. The braking distance (m) is given by:
\[
d_b = 0.039 \times \frac{V^2}{a},
\]
with \( a = 3.4\, \text{m/s}^2 \), the recommended deceleration rate. The total stopping sight distance is then:
\[
\text{SSD} = d_{\text{PRT}} + d_b.
\]

Unlike conventional road design, where SSD is based on design speed, we compute SSD dynamically using the vehicle's actual speed at the time of incident detection. This reflects real-world traffic conditions, where speed varies due to congestion, signals, and geometry. In the simulation, vehicles begin to decelerate once their distance to the incident is less than or equal to the computed SSD.

To model vehicle behavior near the incident, we incorporate principles from Move Over laws, which require drivers to reduce speed or change lanes near stationary roadside vehicles \citep{TrafficI29:online}. We implement a conservative reduced speed of 5 mph (approximately 8 km/h) when a vehicle is within SSD of the incident, representing cautious traversal of the affected edge.

\subsubsection{Contextual lane-changing}

In T-REX, we enhance the realism of lane-changing behavior for vehicles affected by incidents by modifying SUMO’s LC2013 model \citep{erdmann2015sumo} via TraCI. Specifically, vehicles are modeled to make early, strategic lane changes in anticipation of downstream obstructions, reflecting efforts to maintain route adherence and avoid blocked lanes. The model also accounts for speed-seeking behavior, where drivers shift to adjacent lanes perceived to offer better flow conditions. Additionally, cooperative interactions are included, allowing vehicles in neighboring lanes to yield space for merging vehicles when appropriate. The preference for keeping to specific lanes (e.g., the rightmost lane) is relaxed to prioritize incident avoidance. These adaptations are dynamically triggered based on a vehicle’s proximity to an incident: lane-changing parameters are modified when a vehicle enters the SSD range of a disabled vehicle and revert to default once the vehicle has passed. This enables more anticipatory and context-aware lane-changing behavior during disruptions. Detailed settings for the lane-changing model are presented in Appendix \ref{secA1}.


\subsection{Interacting with RL-TSC agents}

As discussed in Section 2.2, this study presents a specific application of T-REX as an environment for training and evaluating RL-TSC methods in the presence of traffic incidents. To enable such learning, it is essential to have a framework that allows RL agents to interact with the traffic simulation—observing states, taking actions, and receiving rewards. In this work, we adopt RESCO (REinforced Signal COntrol) \citep{ault2021reinforcement} to facilitate communication between the simulation environment and RL-TSC algorithms. RESCO is a benchmarking toolkit developed on top of SUMO and integrated with the OpenAI Gym interface, enabling standardized training and evaluation of RL-based control strategies. Specifically, at predefined simulation intervals, RESCO allows the RL agent to receive traffic-related observations via TraCI commands. Based on these observations, the agent selects an action, which is then applied through TraCI, after which the simulation advances by a fixed number of steps. The agent subsequently receives a reward based on traffic performance metrics, observes the new state, and updates its policy accordingly. This interactive loop is repeated over multiple episodes, allowing the agent to progressively learn and refine effective traffic signal control strategies under incident scenarios.

\section{Computational Experiment}
\label{sec4:experiment}

\subsection{Traffic networks}

\begin{figure}[H]
     \centering
     \begin{subfigure}[b]{0.3\columnwidth}
         \centering
         \includegraphics[width=\columnwidth]{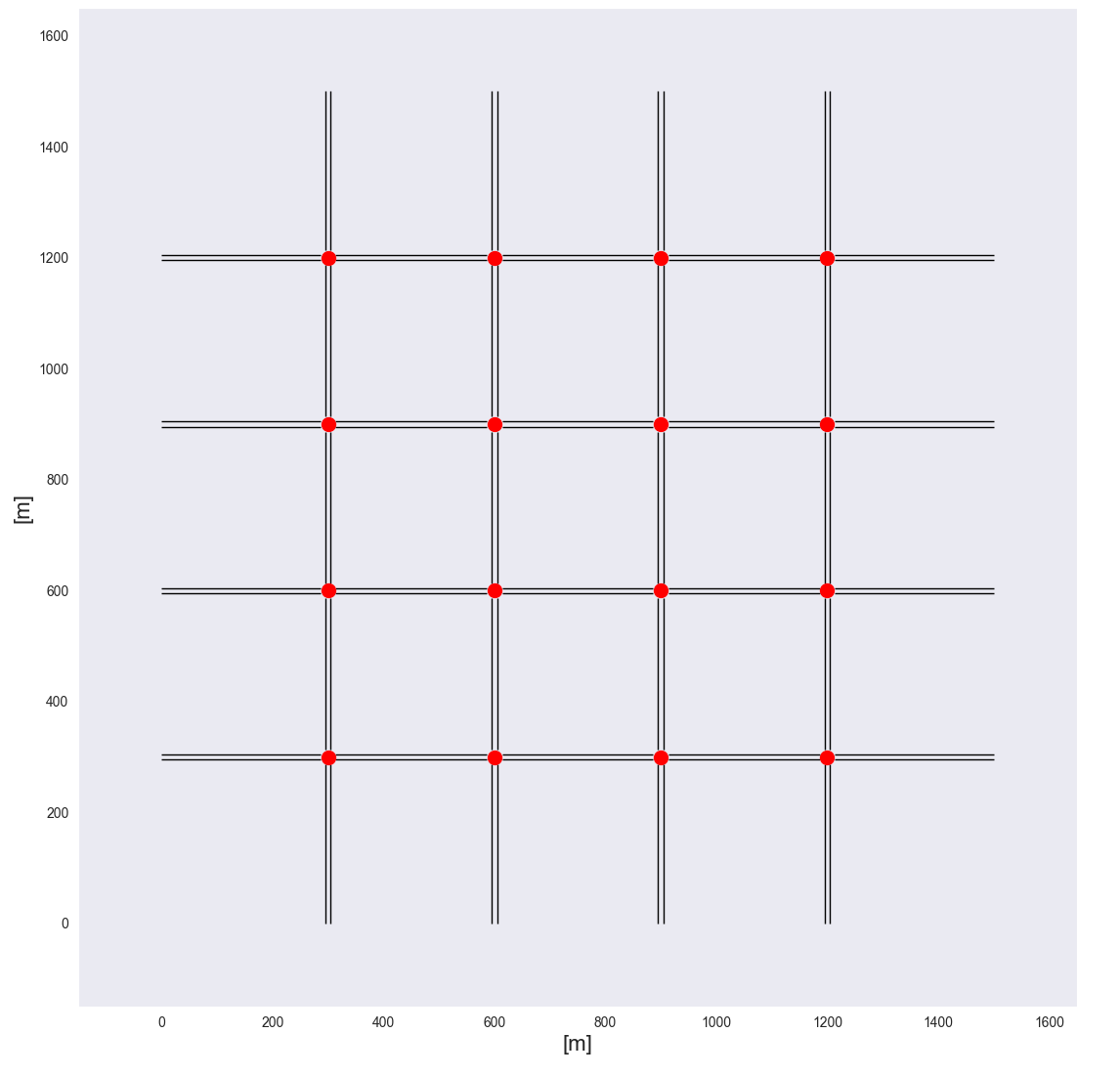}
         \caption{Grid network $4 \times 4$}
         \label{f21}
     \end{subfigure}
     \hfill
     \begin{subfigure}[b]{0.3\columnwidth}
         \centering
         \includegraphics[width=\columnwidth]{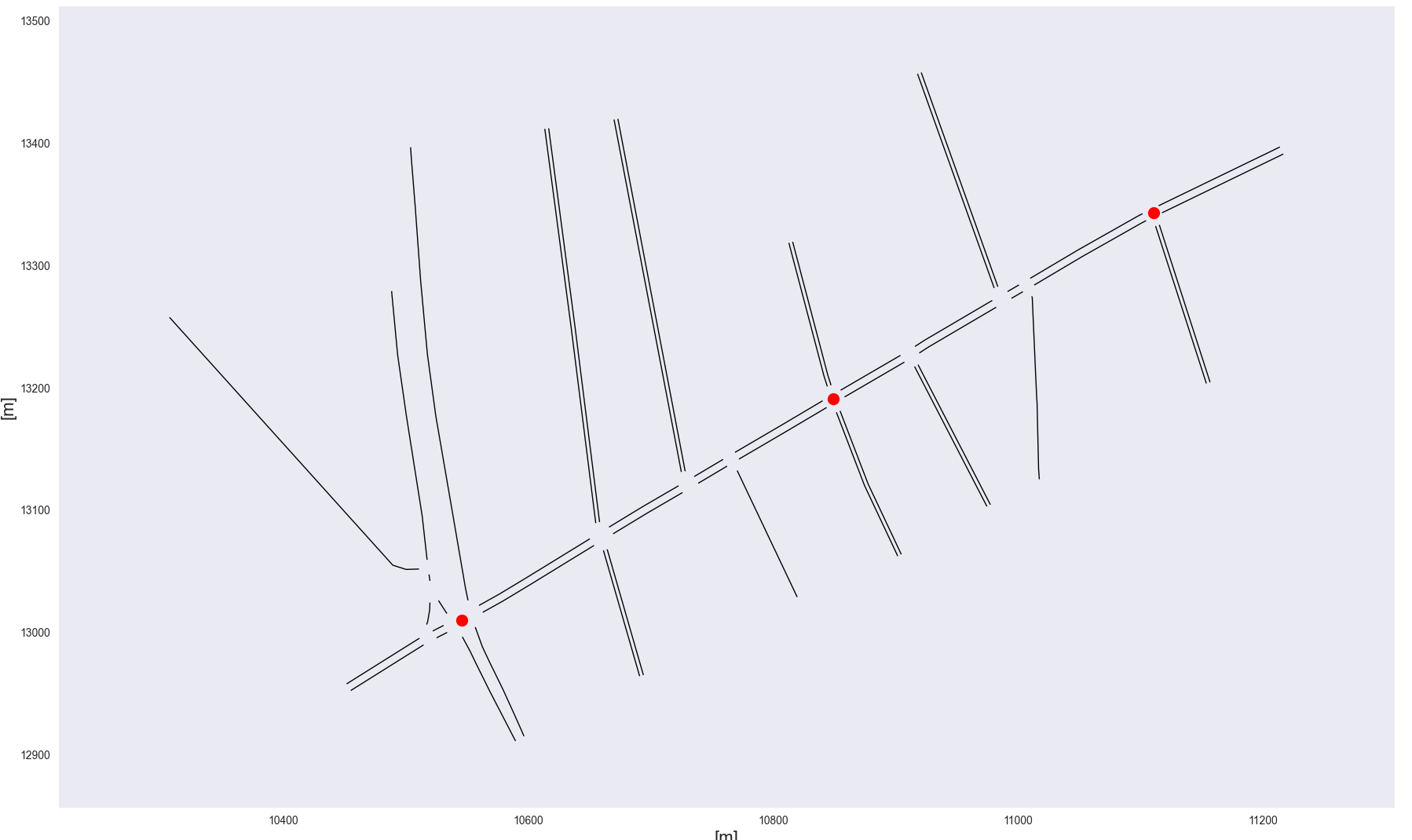}
         \caption{Cologne corridor}
         \label{f22}
     \end{subfigure}
     \hfill
     \begin{subfigure}[b]{0.3\columnwidth}
         \centering
         \includegraphics[width=\columnwidth]{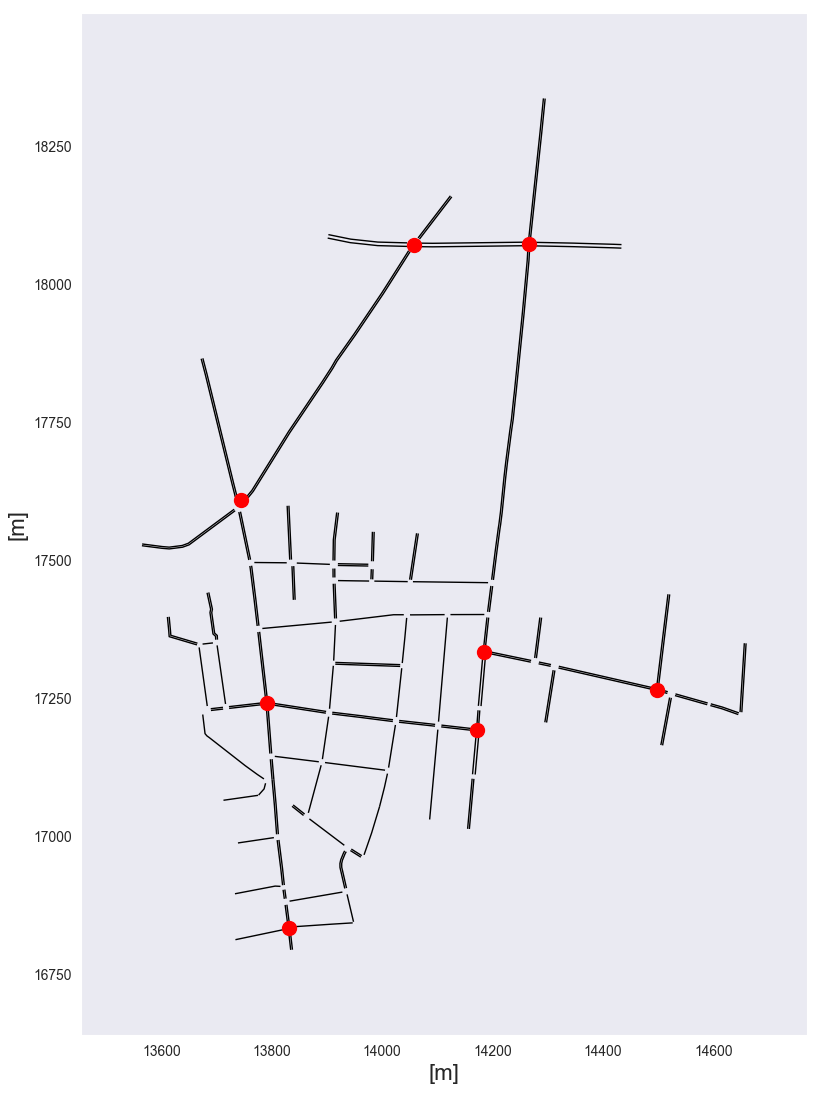}
         \caption{Cologne region}
         \label{f23}
     \end{subfigure}
     \begin{subfigure}[b]{0.3\columnwidth}
         \centering
         \includegraphics[width=\columnwidth]{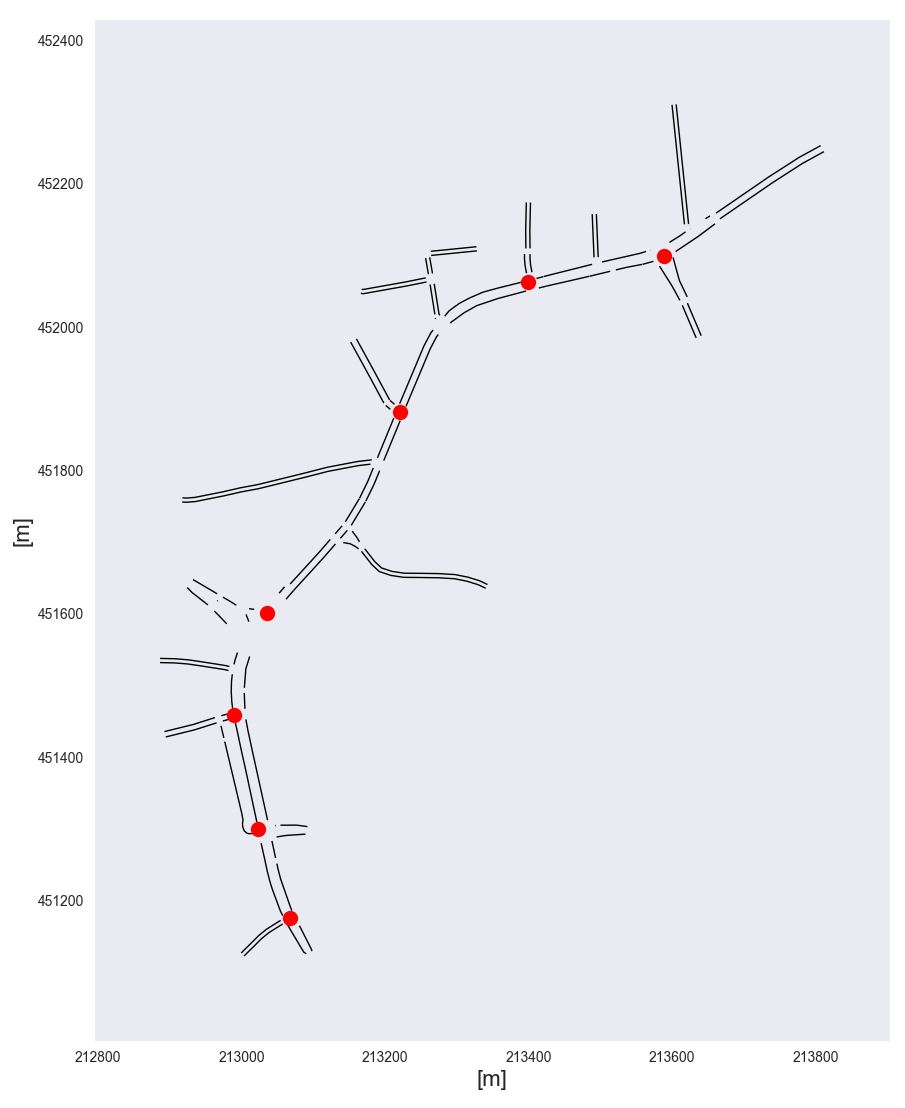}
         \caption{Ingolstadt corridor}
         \label{f24}
     \end{subfigure}
     \begin{subfigure}[b]{0.3\columnwidth}
         \centering
         \includegraphics[width=\columnwidth]{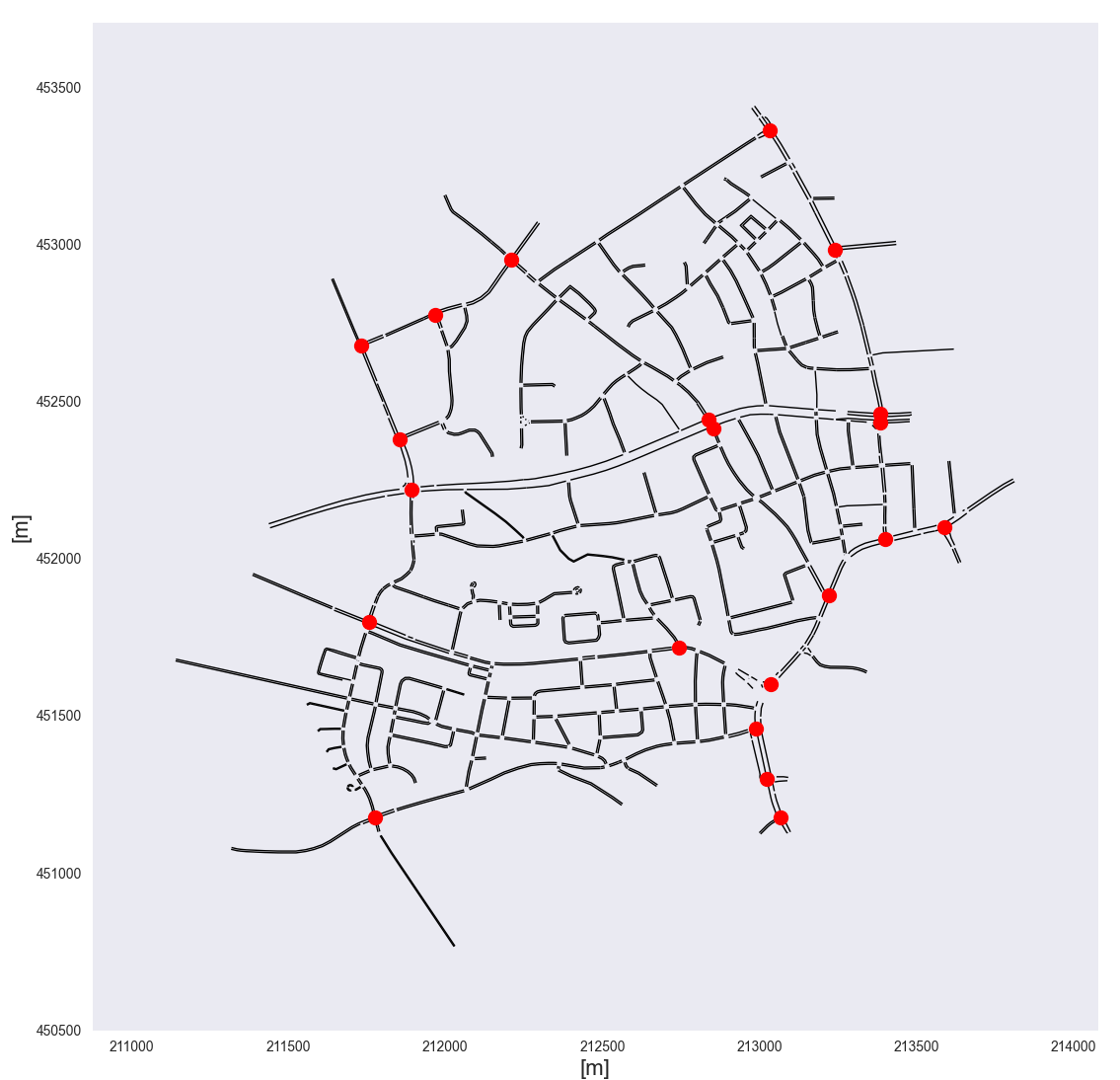}
         \caption{Ingolstadt region}
         \label{f25}
     \end{subfigure}
        \caption{Benchmark scenarios}
        \label{F2}
\end{figure}

To evaluate the robustness of RL-TSC methods at the network level using T-REX, we focus exclusively on multi-intersection traffic networks. Specifically, our experiments are conducted on one synthetic network and four real-world networks. The synthetic grid network, adopted from \cite{chen2020toward}, is a \(4 \times 4\) intersection layout, as illustrated in Figure~\ref{f21}. The real-world network scenarios are based on those used in the experiments of \cite{ault2021reinforcement}, which draw from two well-established SUMO benchmark scenarios: TAPAS Cologne \citep{varschen2006mikroskopische} and InTAS \citep{lobo2020intas}. From TAPAS Cologne, we use two configurations: the Cologne corridor, consisting of 3 intersections (Figure~\ref{f22}), and the Cologne region, which comprises eight intersections (Figure~\ref{f23}). From InTAS, we use the Ingolstadt corridor, with seven intersections (Figure~\ref{f24}), and the Ingolstadt region, which includes 21 intersections (Figure~\ref{f25}). These scenarios represent realistic urban traffic networks in two German cities—Cologne and Ingolstadt—featuring actual road layouts and calibrated traffic demand. Their inclusion enables us to benchmark the performance of RL-TSC methods under realistic conditions and compare them to results from prior studies conducted in non-incident scenarios.

\subsection{RL-based traffic signal control methods and baselines}

We evaluate a diverse set of RL-TSC methods, covering both independent and coordinated learning paradigms for network-level control. Each intersection is modeled as an agent, with decentralized training preferred due to scalability constraints. The selected RL-TSC methods include:

\begin{itemize}
    \item Independent Deep Q-Network (IDQN) is a multi-agent RL method in which each agent independently controls a single intersection without sharing information with other agents. Convolutional layers are used to aggregate lane-level traffic state features, as described in \cite{ault2019learning}.
    \item Independent Proximal Policy Optimization (IPPO) follows the same decentralized architecture as IDQN but utilizes a policy gradient approach (PPO) instead of value-based Q-learning. Including IPPO enables a comparative analysis between value-based and policy gradient methods under incident scenarios—two foundational paradigms in reinforcement learning.

    \item MPLight \citep{chen2020toward} is a decentralized RL method that employs pressure-based coordination. It incorporates queue pressure into the agent’s state representation, allowing partial observability of neighboring intersections. MPLight calculates Q-values for each signal phase based on the phase competition concept derived from the FRAP architecture \citep{zheng2019learning}. Parameter sharing across agents also enhances its scalability in large-scale traffic networks.

    \item Feudal Multi-agent Actor-Critic (FMA2C) \cite{ma2020feudal} extends the state-of-the-art MA2C framework \citep{chu2019multi} by introducing hierarchical coordination via feudal reinforcement learning. In this architecture, a set of intersection-level agents (workers) operate under the guidance of regional agents (managers), enabling coordination through both local learning and top-down goal setting.
\end{itemize}

These methods are chosen for their diversity in learning strategies and coordination levels. All have been validated under non-incident settings \citep{ault2021reinforcement, ault2019learning, ma2020feudal, chen2020toward}, and we adopt their open-source implementations for consistency. Further details on the MDP of each RL-TSC method are presented in Appendix \ref{A4}. 

We also include four traditional rule-based baselines:

\begin{itemize}
    \item Fixed-time control, where phase ordering and durations follow a predefined cycle, based either on real-world traffic signal plans or existing benchmark settings (e.g., for the $4 \times 4$ grid from \cite{chen2020toward}).
    \item Random control, which randomly selects a signal phase at each decision step, providing a naive baseline for comparison.
    \item Max-pressure control, similar to the implementation in \cite{chen2020toward}, selects the phase with the maximum pressure—defined as the difference between upstream and downstream queue lengths.
    \item Greedy control, as used in \cite{ma2020feudal}, selects the phase with the highest wave, where the wave is measured by the number of approaching vehicles at the intersection.
\end{itemize}

These rule-based baselines serve as interpretable and practical benchmarks for evaluating the effectiveness of RL-based methods. They help determine whether the added complexity of RL is justified by improved performance and robustness. Furthermore, comparing RL methods against these baselines under incident scenarios allows us to assess where RL methods provide tangible advantages—such as adaptability to dynamic traffic patterns and resilience to unexpected disruptions.

\subsection{Evaluation metrics}
\label{sec3:metrics}

Previous studies on RL-TSC have primarily focused on conventional traffic metrics such as queue length, travel time, and delay. While RL-TSC methods often outperform traditional approaches on these metrics, the comparisons are not always aligned with real-world deployment challenges \citep{han2023leveraging}. Notably, many studies report best-case performance during training, which may not reflect reliability or robustness. Strong simulation results do not guarantee real-world effectiveness, especially for RL algorithms that rely heavily on exploration and are often sample inefficient.

In practice, the real-world deployment of RL-TSC can follow three main approaches: pre-training (training fully in simulation), online learning (learning directly and exclusively in the real world), and pre-training with continuous learning (fine-tuning a pre-trained policy using real-time data). For online and hybrid approaches, the learning phase after deployment is critical, as real-time adaptation affects both traffic efficiency and safety.

Evaluating the robustness of learning requires assessing learning stability, convergence behavior, and the cost of exploration—particularly the degradation caused by random actions during early training. To systematically quantify these aspects, we propose the following novel metrics:

\textbf{Learning Stability Index (LSI)} measures the variance in learning performance across episodes, indicating how stable the training process is in uncertain environments:
\[
LSI = \frac{1}{T} \sum_{t=1}^{T} (I_t - I_{t-1})^2,
\]
where \(I_t\) is the performance indicator at episode \(t\), and \(T\) is the total number of episodes.

\textbf{Final Performance Deviation (FPD)} captures the difference between final and best observed performance. Since DRL methods often exhibit unstable convergence \citep{henderson2018deep}, this metric reflects policy retention:
\[
FPD = \frac{|I_{\text{final}} - I_{\text{best}}|}{I_{\text{best}}},
\]
where \(I_{\text{final}}\) and \(I_{\text{best}}\) are the final and peak performance values during training.

\textbf{Convergence Rate (CR)} assesses how quickly performance stabilizes:
\[
CR = \begin{cases}
\frac{1}{T_c}, & \text{if } I_{\text{final}} < I_0 \\
-\frac{1}{T_c}, & \text{if } I_{\text{final}} > I_0 \\
0, & \text{otherwise}
\end{cases}
\]
where \(I_0\) is initial performance, \(I_{\text{final}}\) is final performance, and \(T_c\) is the convergence time (i.e., when performance stays within a threshold \(\epsilon\) of the final value). Negative CR indicates divergence.

\textbf{Area Under the Learning Curve (AUC)} evaluates overall training efficiency by integrating performance over time:
\[
AUC = \int_0^T I(t) \, dt,
\]
where lower AUC implies faster and more efficient convergence.

To assess robustness under disruptions, we compare AUC values in normal and perturbed conditions using the \textbf{Relative AUC Difference (RAUC)}:
\[
RAUC = \frac{AUC_{G'} - AUC_G}{AUC_G},
\]
where \(AUC_G\) is from the original environment and \(AUC_{G'}\) from the perturbed one. Lower RAUC suggests better robustness to incidents.

In addition to training performance, testing and generalization are crucial for real-world deployment, especially when using pre-training or hybrid learning. Discrepancies between simulated and real-world environments often lead to performance degradation during transfer \citep{han2023leveraging}. To quantify this, we define the \textbf{Performance Degradation Index} (PDI) as:
\[
PDI = \frac{I_{G'} - I_G}{I_G} = \frac{\Delta I_G}{I_G},
\]
where \(I_G\) is the performance in the training environment \(G\), and \(I_{G'}\) is performance in the testing environment \(G'\). A higher PDI indicates greater degradation and lower generalization capability.

\subsection{Experimental Setup}

Building on the robustness-oriented evaluation metrics, we conduct a series of experiments to assess the performance of RL-TSC methods across multiple dimensions of robustness. These dimensions—learning performance, generalization, and transferability/adaptation—correspond to key evaluation aspects that are critical under different real-world deployment strategies of RL-TSC. Across all experiments, we consider multiple traffic networks and include non-learning baselines for benchmarking. Each traffic episode simulates 3,600 seconds, including a 100-second warm-up phase.

\textbf{Experiment 1: Learning performance.}  
In this experiment, we evaluate the learning performance of four RL-TSC methods—IDQN, MPLight, IPPO, and FMA2C—under both normal and incident scenarios across various traffic networks. While simulation-based results may not directly translate to real-world outcomes, they provide valuable insights into the learning behavior and convergence characteristics of each method, particularly under deployment strategies involving online training or continuous adaptation.

Value-based methods (IDQN, MPLight) are trained for 100 episodes, while policy-gradient methods (IPPO, FMA2C), which generally require more extensive training, are trained for 1,400 episodes. Training is repeated five times using different random seeds for statistical robustness. Performance is measured using average queue length, travel time, waiting time, and delay. For evaluation, we report the average over the top-performing episodes—specifically, the best 10 consecutive episodes for IDQN and MPLight, and the best 100 consecutive episodes for IPPO and FMA2C. Incident scenarios include two randomly generated disruptions per episode to induce congestion propagation, rerouting, and conflicting demand patterns.

\textbf{Experiment 2: Testing performance and generalization.}  
This experiment investigates the ability of RL-TSC methods to generalize across environments without additional training—a scenario representative of pre-training deployment where real-world data is limited. We evaluate three methods—IDQN, MPLight, and FMA2C—by training them in one scenario and testing them in another without further adaptation (i.e., zero-shot generalization). Due to overfitting observed during training, IPPO is excluded from this experiment.

Specifically, we examine four training–testing combinations: (i) training and testing in the base (no-incident) scenario, (ii) training in the base scenario and testing in the incident scenario, (iii) training and testing in the incident scenario, and (iv) training in the incident scenario and testing in the base scenario. For each method, we select the best policy based on the lowest average travel time over ten consecutive training episodes and evaluate it directly in the target test scenario.

\textbf{Experiment 3: Transferability and online adaptation.}  
This experiment evaluates how well RL-TSC methods adapt to dynamic conditions under a deployment strategy that combines pre-training with continued online learning. Each method is first trained in a base scenario without incidents and then transferred to an incident scenario, where training continues without model reinitialization.

IDQN and MPLight are trained for 100 episodes per phase (pre-transfer and post-transfer), while FMA2C is trained for 1,400 episodes per phase to account for its greater training complexity. The experiments are conducted on both the synthetic Grid 4×4 network and the real-world Ingolstadt Region network. We assess adaptation and resilience by observing changes in average travel time across the training and transfer phases.

\section{Results and discussion}
\label{sec5:results}

\subsection{Experiment 1: Learning performance}

Table \ref{tab1:results} presents the best performance achieved during training for both the base (normal) and incident scenarios. Bolded values indicate the best-performing RL-TSC method for each metric. Using travel time as the primary performance indicator, we compute the robustness metrics defined in Section \ref{sec3:metrics}, with the results summarized in Table \ref{tab2:robustness}. These metrics include LSI, FPD, CR, and RAUC. For all metrics, lower values indicate better performance.

\begin{table}
\caption{Performance of the investigated methods on training}
\label{tab1:results}
\begin{tabular}{@{}lcccccccc@{}}
\toprule
\multirow{2}{*}{\textbf{Grid}} & \multicolumn{2}{c}{\textbf{Avg. Queue}} & \multicolumn{2}{c}{\textbf{Avg. Travel time}} & \multicolumn{2}{c}{\textbf{Avg. Waiting}} & \multicolumn{2}{c}{\textbf{Avg. Delay}} \\
                               & \textbf{Base}    & \textbf{Incident}    & \textbf{Base}        & \textbf{Incident}      & \textbf{Base}      & \textbf{Incident}    & \textbf{Base}     & \textbf{Incident}   \\
\midrule
Fixed-time                     & -                & -                    & 204.51               & 214.07                 & 65.09              & 73.01                & 93.26             & 103.23              \\
Random                         & 2.56             & 2.85                 & 242.67               & 255.75                 & 101.09             & 112.23               & 132.18            & 144.89              \\
Max-pressure                    & 0.61             & 1.09                 & 160.14               & 181.75                 & 23.11              & 42.50                & 49.06             & 70.27               \\
Greedy                         & 0.30             & 0.39                 & 145.41               & 160.78                 & 10.96              & 24.28                & 34.51             & 49.42               \\
IDQN                           & \textbf{0.35}    & 1.25                 & \textbf{145.79}      & 191.92                 & \textbf{12.49}     & 51.27                & \textbf{34.25}    & 80.72               \\
MPLight                        & 0.60             & \textbf{1.14}        & 160.38               & \textbf{187.01}        & 21.97              & \textbf{45.77}       & 49.06             & \textbf{75.27}      \\
IPPO                           & 2.11             & 3.25                 & 223.95               & 271.16                 & 84.49              & 128.70               & 113.34            & 161.01              \\
FMA2C                          & 1.94             & 2.32                 & 215.75               & 234.10                 & 76.38              & 92.69                & 105.07            & 122.98              \\
\midrule
\textbf{Cologne Cor.}          & \textbf{}        & \textbf{}            & \textbf{}            & \textbf{}              & \textbf{}          & \textbf{}            & \textbf{}         & \textbf{}           \\
\midrule
Fixed-time                     & -                & -                    & 76.23                & 116.94                 & 24.89              & 62.88                & 38.85             & 79.64               \\
Random                         & 6.96             & 9.00                 & 140.65               & 203.31                 & 82.85              & 142.43               & 124.22            & 203.68              \\
Max-pressure                    & 4.38             & 7.95                 & 128.94               & 190.10                 & 78.98              & 138.62               & 95.32             & 165.51              \\
Greedy                         & 2.53             & 3.16                 & 135.07               & 157.74                 & 87.04              & 106.71               & 101.16            & 139.03              \\
IDQN                           & \textbf{2.48}    & \textbf{4.81}        & 91.60                & 150.16                 & 38.13              & 95.99                & 61.69             & 132.75              \\
MPLight                        & 5.11             & 6.46                 & 111.00               & 178.74                 & 54.87              & 124.70               & 89.84             & 169.90              \\
IPPO                           & 4.20             & 7.30                 & 114.53               & 189.66                 & 62.04              & 134.47               & 82.41             & 172.51              \\
FMA2C                          & 2.89             & 4.85                 & \textbf{88.35}       & \textbf{126.96}        & \textbf{37.30}     & \textbf{73.61}       & \textbf{53.92}    & \textbf{101.58}     \\
\midrule
\textbf{Cologne Reg.}          & \textbf{}        & \textbf{}            & \textbf{}            & \textbf{}              & \textbf{}          & \textbf{}            & \textbf{}         & \textbf{}           \\
\midrule
Fixed-time                     & -                & -                    & 114.51               & 128.91                 & 29.37              & 41.20                & 49.37             & 62.34               \\
Random                         & 4.69             & 4.81                 & 166.97               & 177.91                 & 62.32              & 70.77                & 102.30            & 115.03              \\
Max-pressure                    & 0.61             & 0.98                 & 91.77                & 107.97                 & 8.63               & 22.81                & 27.63             & 44.08               \\
Greedy                         & 0.30             & 0.34                 & 84.89                & 93.29                  & 4.69               & 11.65                & 20.80             & 29.72               \\
IDQN                           & \textbf{0.39}    & \textbf{0.81}        & \textbf{86.37}       & \textbf{101.60}        & \textbf{5.96}      & \textbf{17.50}       & \textbf{22.34}    & \textbf{37.58}      \\
MPLight                        & 1.01             & 1.10                 & 99.27                & 107.34                 & 13.98              & 20.50                & 35.12             & 42.88               \\
IPPO                           & 1.25             & 4.59                 & 110.78               & 178.36                 & 24.98              & 74.33                & 46.92             & 115.07              \\
FMA2C                          & 1.02             & 1.29                 & 98.12                & 109.07                 & 14.76              & 22.69                & 33.86             & 44.86               \\
\midrule
\textbf{Ingolstadt Cor.}       & \textbf{}        & \textbf{}            & \textbf{}            & \textbf{}              & \textbf{}          & \textbf{}            & \textbf{}         & \textbf{}           \\
\midrule
Fixed-time                     & -                & -                    & 116.91               & 175.84                 & 48.07              & 101.25               & 73.01             & 131.62              \\
Random                         & 4.42             & 5.28                 & 126.44               & 156.56                 & 52.15              & 79.54                & 94.50             & 135.97              \\
Greedy                         & 1.35             & 2.23                 & 79.79                & 111.11                 & 15.28              & 44.39                & 42.23             & 88.75               \\
Max-pressure                    & 1.81             & 2.19                 & 100.58               & 120.19                 & 34.60              & 53.10                & 77.38             & 105.83              \\
Greedy                         & \textbf{0.74}    & \textbf{1.97}        & \textbf{72.43}       & \textbf{103.34}        & \textbf{10.22}     & \textbf{37.02}       & \textbf{35.29}    & \textbf{75.94}      \\
MPLight                        & 1.39             & 2.95                 & 78.78                & 125.89                 & 16.53              & 59.54                & 46.43             & 108.18              \\
IPPO                           & 1.29             & 3.71                 & 104.19               & 132.52                 & 38.48              & 61.06                & 69.12             & 110.21              \\
FMA2C                          & 1.88             & 2.73                 & 87.13                & 113.51                 & 22.48              & 45.42                & 53.27             & 89.42               \\
\midrule
\textbf{Ingolstadt Reg.}       & \textbf{}        & \textbf{}            & \textbf{}            & \textbf{}              & \textbf{}          & \textbf{}            & \textbf{}         & \textbf{}           \\
\midrule
Fixed-time                     & -                & -                    & 295.22               & 309.06                 & 102.41             & 114.40               & 150.12            & 163.15              \\
Random                         & 4.46             & 4.61                 & 326.63               & 349.36                 & 124.66             & 146.50               & 192.79            & 218.46              \\
Max-pressure                    & 5.40             & 5.53                 & 551.37               & 550.81                 & 358.23             & 357.09               & 420.98            & 424.40              \\
Greedy                         & 2.26             & 2.50                 & 301.70               & 334.42                 & 114.37             & 142.79               & 159.59            & 192.27              \\
IDQN                           & \textbf{1.67}    & \textbf{1.91}        & \textbf{242.88}      & \textbf{263.05}        & \textbf{60.72}     & \textbf{76.27}       & \textbf{106.78}   & \textbf{127.88}     \\
MPLight                        & 3.14             & 3.36                 & 284.65               & 306.57                 & 92.14              & 110.18               & 147.40            & 169.86              \\
IPPO                           & 3.54             & 3.99                 & 310.40               & 337.09                 & 115.26             & 142.31               & 176.84            & 204.88              \\
FMA2C                          & 2.15             & 2.60                 & 252.65               & 281.96                 & 65.12              & 90.40                & 117.15            & 147.93       \\
\bottomrule
\end{tabular}
\end{table}

Under normal conditions, RL-TSC methods generally outperform traditional baselines. For instance, in the Ingolstadt Region, IDQN—representing independent value-based learning—achieves 242.88 seconds in travel time, outperforming Fixed-Time (295.22s). This suggests that even without inter-agent coordination, well-structured local state representations and value-based updates can yield effective control. Meanwhile, FMA2C, which leverages hierarchical coordination through feudal learning, performs best in the Cologne Corridor (88.35s), showing that coordinated decision-making across layers enhances optimization in more complex urban environments.

However, this advantage does not consistently carry over to incident scenarios. In the Cologne Corridor, MPLight—representing decentralized coordination using pressure-based reward design—experiences significant performance degradation, with travel time increasing to 178.74 seconds, falling behind Greedy (157.74s) and Fixed-Time (116.94s). This suggests that while pressure-based mechanisms can facilitate implicit coordination under stable conditions, they may be less robust to sudden disruptions. Similarly, IPPO, as a representative of decentralized policy-gradient methods, underperforms Greedy in the Ingolstadt Corridor (132.52s vs. 120.19s in travel time), indicating that its slower convergence and higher variance may hinder responsiveness to incident-induced dynamics.

In the synthetic Grid4x4 network, all RL-TSC methods underperform relative to the Greedy baseline during incidents. For example, IDQN’s travel time rises from 145.79 to 191.92 seconds, while Greedy increases only modestly from 145.41 to 160.78 seconds. This performance gap can be attributed to the grid's uniform and predictable topology, where simple heuristics suffice. In contrast, irregular real-world networks such as Ingolstadt offer more opportunities for adaptive RL policies to outperform rule-based methods.

Robustness metrics reveal significant degradation in learning performance under incidents. IDQN, as an independent learner, suffers from the most volatile behavior: its LSI in Grid4x4 spikes from 7.33 to 814.66, and its RAUC reaches 24.06\%, indicating instability in adapting to dynamic disruptions. IPPO exhibits a similar pattern, with a RAUC of 417.55\% in the Ingolstadt Corridor, underscoring the challenges of sample inefficiency and sensitivity in policy gradient methods. Negative CR values, such as IDQN’s drop from 0.059 to –0.013 in the Cologne Corridor, further indicate convergence failure in turbulent environments. These issues are also reflected in MPLight’s performance, highlighting limitations of pressure-based coordination when facing non-stationary disruptions.

Among all evaluated methods, FMA2C consistently demonstrates the highest robustness. It maintains low LSI and FPD values and achieves the lowest RAUC in several networks—for example, only 9.38\% in the Ingolstadt Region. Its hierarchical architecture allows for structured cooperation: high-level agents adapt strategy while low-level agents manage local actions, providing resilience to large-scale disturbances. However, this robustness is offset by slower convergence and sample inefficiency. For instance, FMA2C’s CR values remain low across networks (e.g., 0.001 in the Cologne Region), reflecting the cost of learning coordinated behavior. This trade-off suggests that while hierarchical RL is well-suited for complex, disruption-prone networks, its use in real-time systems may be limited by the need for random exploration and sample inefficiency.

\begin{sidewaystable}[]
\caption{Learning performance metrics of RL-TSC methods}
\label{tab2:robustness}
\begin{tabular}{lccccccccc}
\toprule
\multirow{2}{*}{\textbf{Grid4x4}} & \multicolumn{2}{c}{\textbf{LSI}}  & \multicolumn{2}{c}{\textbf{FPD}}  & \multicolumn{2}{c}{\textbf{CR}}   & \multicolumn{2}{c}{\textbf{AUC}}  & \multirow{2}{*}{\textbf{RAUC}} \\
                                  & \textbf{Base} & \textbf{Incident} & \textbf{Base} & \textbf{Incident} & \textbf{Base} & \textbf{Incident} & \textbf{Base} & \textbf{Incident} &                                \\
                                  \midrule
IDQN                              & 7.330         & 814.661           & 0.034         & 0.342             & 0.014         & 0.067             & 17050.706     & 21152.862         & 24.059                         \\
MPLight                           & 7.486         & 612.472           & 0.007         & 0.346             & 0.013         & 0.080             & 17979.397     & 21173.720         & 17.767                         \\
IPPO                              & 875.889       & 3001.635          & 2.174         & 4.770             & -0.0008       & -0.001            & 665797.802    & 1241386.056       & 86.451                         \\
FMA2C                             & 19.838        & 330.332           & 0.034         & 0.092             & 0.002         & 0.042             & 318017.798    & 351935.840        & 10.665                         \\
\midrule
\textbf{Cologne Cor.}                 & \textbf{}     & \textbf{}         & \textbf{}     & \textbf{}         & \textbf{}     & \textbf{}         & \textbf{}     & \textbf{}         &                                \\
\midrule
IDQN                              & 7775.667      & 114.436           & 1.543         & 1.179             & 0.059         & -0.013            & 13626.695     & 22531.990         & 65.352                         \\
MPLight                           & 5695.477      & 140.097           & 1.373         & 1.257             & -0.014        & -0.012            & 17114.030     & 24729.504         & 44.498                         \\
IPPO                              & 10657.108     & 12242.250         & 9.966         & 3.621             & -0.007        & -0.004            & 365325.553    & 219815.749        & -39.830                        \\
FMA2C                             & 8212.069      & 9967.432          & 0.200         & 1.621             & 0.002         & 0.023             & 199662.804    & 279639.156        & 40.056                         \\
\midrule
\textbf{Cologne Reg.}                 & \textbf{}     & \textbf{}         & \textbf{}     & \textbf{}         & \textbf{}     & \textbf{}         & \textbf{}     & \textbf{}         &                                \\
\midrule
IDQN                              & 51.853        & 656.083           & 0.009         & 0.169             & 0.013         & 0.029             & 10278.780     & 12223.445         & 18.919                         \\
MPLight                           & 511.220       & 676.036           & 1.435         & 1.173             & -0.010        & -0.010            & 13185.565     & 13442.965         & 1.952                          \\
IPPO                              & 112.498       & 1711.564          & 0.192         & 2.045             & 0.031         & -0.002            & 183600.283    & 528496.517        & 187.852                        \\
FMA2C                             & 90.924        & 516.186           & 0.020         & 0.024             & 0.001         & 0.001             & 164846.319    & 190315.011        & 15.450                         \\
\midrule
\textbf{Ingolstadt Cor.}              & \textbf{}     & \textbf{}         & \textbf{}     & \textbf{}         & \textbf{}     & \textbf{}         & \textbf{}     & \textbf{}         &                                \\
\midrule
IDQN                              & 26.754        & 1840.784          & 0.017         & 0.579             & 0.013         & 0.333             & 8676.864      & 13636.753         & 57.162                         \\
MPLight                           & 104.527       & 5903.952          & 0.040         & 2.457             & 0.015         & -0.010            & 9182.465      & 19714.950         & 114.702                        \\
IPPO                              & 62.338        & 4578.024          & 0.060         & 5.111             & 0.033         & -0.002            & 151446.714    & 783816.170        & 417.552                        \\
FMA2C                             & 99.902        & 2152.070          & 0.030         & 0.065             & 0.001         & 0.002             & 140259.031    & 205025.106        & 46.176                         \\
\midrule
\textbf{Ingolstadt Reg.}             & \textbf{}     & \textbf{}         & \textbf{}     & \textbf{}         & \textbf{}     & \textbf{}         & \textbf{}     & \textbf{}         &                                \\

\midrule
IDQN                              & 2742.280      & 3805.556          & 0.330         & 0.038             & 0.110         & 0.615             & 26980.021     & 29725.092         & 10.174                         \\
MPLight                           & 6286.990      & 4468.380          & 2.051         & 0.615             & -0.010        & -0.013            & 41709.383     & 37294.750         & -10.584                        \\
IPPO                              & 4725.375      & 4386.292          & 4.140         & 1.716             & -0.001        & -0.002            & 1348850.585   & 1055946.293       & -21.715                        \\
FMA2C                             & 6918.328      & 3211.393          & 0.407         & 0.156             & 0.250         & 0.003             & 427084.614    & 467161.467        & 9.384                      \\
\bottomrule
\end{tabular}
\end{sidewaystable}

\subsection{Experiment 2: Testing performance and generalization}

\begin{figure}
    \centering
    \includegraphics[width=\linewidth]{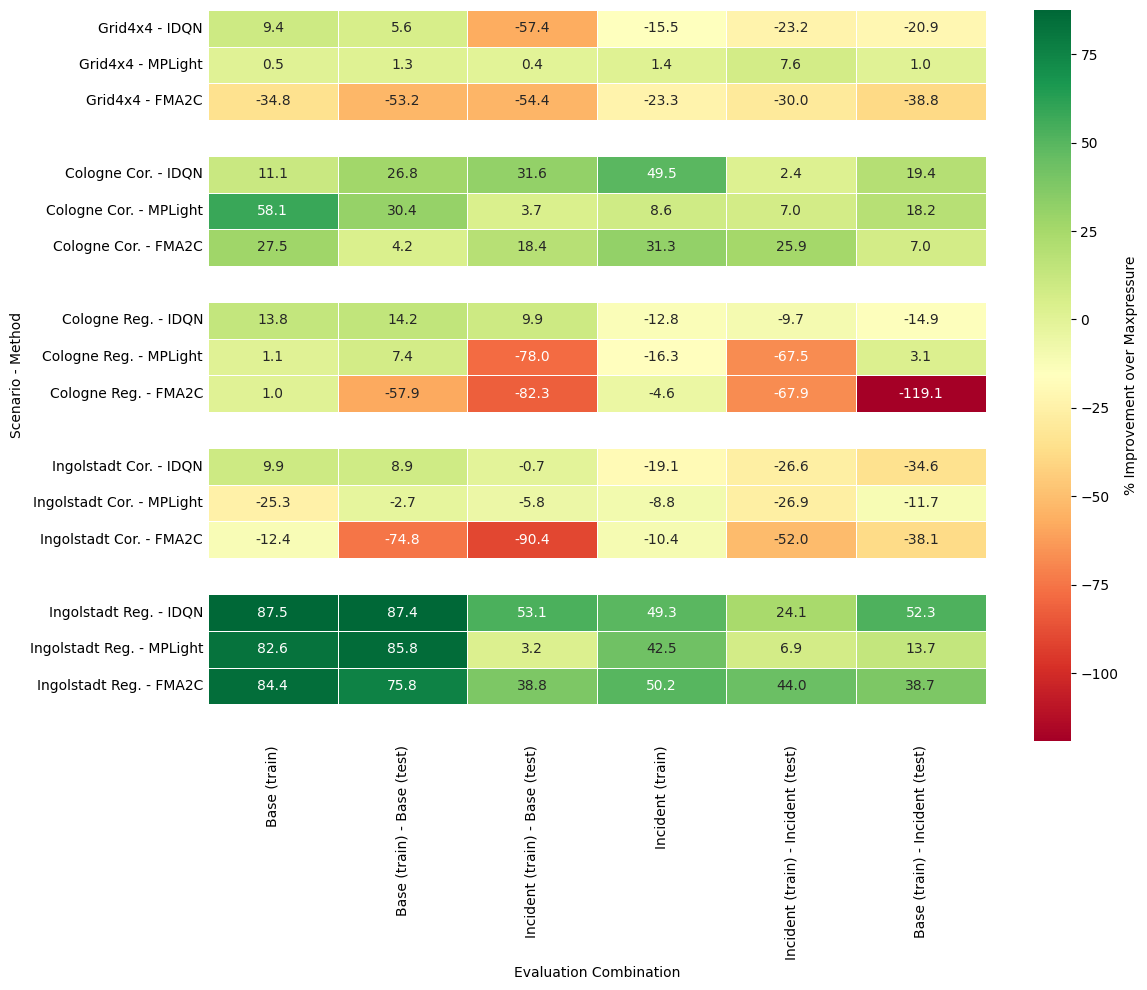}
    \caption{RL-TSC performance improvement over Max-pressure}
    \label{fig:generalization}
\end{figure}

Generalization is quantified using the Performance Degradation Index (PDI), reported in Table~\ref{tab:PDI}, where columns denote training–testing scenario combinations. Detailed performance results are presented in Appendix \ref{A3}. While PDI enables method-level comparison across networks, it is less suited for comparing across training-testing pairs, since scenarios with incidents naturally incur longer travel times. To complement this, we also report each method’s percentage improvement over the Max-pressure baseline under various conditions (Figure~\ref{fig:generalization}).

\begin{table}[]
\centering
\caption{Performance degradation index of RL-TSC methods}
\label{tab:PDI}
\begin{tabular}{lccccc}
\toprule
Grid4x4         & Base-base & Base-incident & Incident-incident & Incident-base & Average \\
\midrule
IDQN            & 0.04      & 0.61          & 0.07              & 0.13          & 0.21    \\
MPLight         & -0.01     & 0.20          & -0.06             & -0.16         & -0.01   \\
FMA2C           & 0.14      & 0.24          & 0.05              & 0.04          & 0.12    \\
\midrule
Cologne Cor.    &           &               &                   &               &         \\
\midrule
IDQN            & -0.18     & 0.36          & 0.93              & -0.09         & 0.25    \\
MPLight         & 0.66      & 1.92          & 0.02              & -0.30         & 0.57    \\
FMA2C           & 0.32      & 0.92          & 0.08              & -0.21         & 0.28    \\
\midrule
Cologne Reg.    &           &               &                   &               &         \\
\midrule

IDQN            & 0.00      & 0.78          & -0.03             & -0.40         & 0.09    \\
MPLight         & -0.06     & 0.31          & 0.44              & 0.15          & 0.21    \\
FMA2C           & 0.60      & 1.95          & 0.61              & 0.31          & 0.87    \\
\midrule
Ingolstadt Cor. &           &               &                   &               &         \\
\midrule
IDQN            & 0.01      & 1.54          & 0.06              & -0.50         & 0.28    \\
MPLight         & -0.18     & 0.52          & 0.17              & -0.43         & 0.02    \\
FMA2C           & 0.55      & 1.09          & 0.38              & 0.01          & 0.51    \\
\midrule
Ingolstadt Reg. &           &               &                   &               &         \\
\midrule
IDQN            & -0.04     & 0.11          & 0.50              & -0.10         & 0.12    \\
MPLight         & 0.63      & 0.65          & 0.62              & 0.64          & 0.63    \\
FMA2C           & 0.41      & 0.40          & 0.12              & 0.20          & 0.28   \\
\bottomrule
\end{tabular}
\end{table}

The observed generalization patterns in our results can be interpreted through the lens of epistemic and aleatoric uncertainty in RL. Epistemic uncertainty, which arises from limited knowledge or insufficient data, reflects the model’s uncertainty about the environment and can often lead to overfitting. Aleatoric uncertainty, in contrast, stems from inherent randomness or noise in the environment and cannot be reduced by collecting more data. Notably, Figure 3 reveals that policies trained in base scenarios in many cases outperform those trained directly under incident conditions when evaluated in the presence of disruptions. This counterintuitive finding can be partially attributed to the epistemic uncertainty inherent in incident-trained models. Since traffic incidents during training vary in location, severity, and timing, the agent may overfit to these specific conditions, leading to brittle policies that fail to generalize to unseen disruptions during testing. This form of model uncertainty, caused by limited and biased exposure to disruptions, limits the agent’s ability to build a robust policy.

In contrast, policies trained in base scenarios may capture more stable and generalizable traffic patterns, allowing them to better handle moderate disruptions during testing—even without direct exposure to such conditions during training. This suggests that reducing epistemic uncertainty through broader or more diverse training experiences may be more beneficial than narrowly focusing on incident-specific conditions.

Additionally, the environment's aleatoric uncertainty—stemming from the inherent stochasticity of traffic behavior, such as variable vehicle arrival times or driver responses—remains irreducible and further challenges generalization. The poor transferability of incident-trained policies to base scenarios also underscores that without explicitly accounting for both types of uncertainty—e.g., through robust algorithmic design, regularization, or uncertainty-aware exploration strategies—RL-TSC methods struggle to maintain performance under train-test distribution shifts.

Method-specific analysis reveals important differences in how RL-TSC approaches handle uncertainty and generalize across scenarios. IDQN, representing the independent value-based learning paradigm, performs reliably when training and testing conditions are aligned (e.g., base–base or incident–incident), but its performance degrades sharply under distribution shifts. In Grid4x4, for example, IDQN improves travel time by 5.62\% over Max-Pressure in the base–base setting but suffers performance drops of 57.41\% and 23.2\% in the incident–base and incident–incident cases, respectively. This sharp decline reflects high epistemic uncertainty—IDQN’s reliance on detailed local state features enables it to model regular traffic dynamics effectively, but it also leads to overfitting to simulation-specific patterns. However, in larger and more complex networks like Ingolstadt, IDQN generalizes better, as indicated by its relatively low PDI values (0.11 for base–incident and 0.50 for incident–incident). This improvement is likely due to the greater structural diversity and rerouting flexibility in such networks, which expose the agent to a broader range of conditions during training and reduce epistemic uncertainty.

MPLight, which follows a decentralized coordination paradigm using pressure-based state and reward representations, exhibits more stable performance across scenarios. Its abstraction over raw traffic features—by focusing on upstream–downstream queue differences—effectively mitigates aleatoric uncertainty by filtering out low-level stochastic variations in traffic flow. In Grid4x4, MPLight maintains consistent performance across all training–testing combinations, achieving the lowest average PDI (–0.01). However, its use of parameter sharing and uniform logic across intersections becomes a limitation in heterogeneous networks like Cologne and Ingolstadt. In these settings, diverse traffic patterns and intersection geometries introduce epistemic uncertainty that MPLight’s generalized design cannot fully capture, leading to higher degradation indices (e.g., 1.92 and 0.65, respectively).

FMA2C, a hierarchical RL-TSC method employing manager–worker coordination, exhibits the most pronounced generalization challenges across multiple settings. In structured, mid-sized networks like Grid4x4 and Cologne Region, FMA2C struggles with epistemic uncertainty arising from the sensitivity of its learned coordination strategies to small shifts in regional traffic flows. This is reflected in consistently high PDI values (e.g., 1.95 in Cologne Region base–incident), indicating disrupted alignment between manager and worker agents. Moreover, the hierarchical structure is also vulnerable to aleatoric uncertainty, as stochastic variations can misalign the intended top-down coordination. Nonetheless, in large-scale, irregular environments such as the Ingolstadt Region, the hierarchical architecture aligns more naturally with real-world traffic patterns, allowing FMA2C to achieve more stable performance (e.g., PDI of 0.12 in incident–incident, 0.20 in incident–base). These results suggest that while hierarchical coordination offers scalability and structure, its effectiveness depends heavily on network characteristics and requires careful adaptation to manage both types of uncertainty.

Our findings highlight the need to better handle both epistemic and aleatoric uncertainties in RL-based TSC, especially when aiming for robust real-world deployment across varying traffic conditions.

\subsection{Experiment 3: Transferability and adaptation}

\begin{figure}
     \begin{subfigure}[b]{0.5\columnwidth}
         \centering
         \includegraphics[width=\columnwidth]{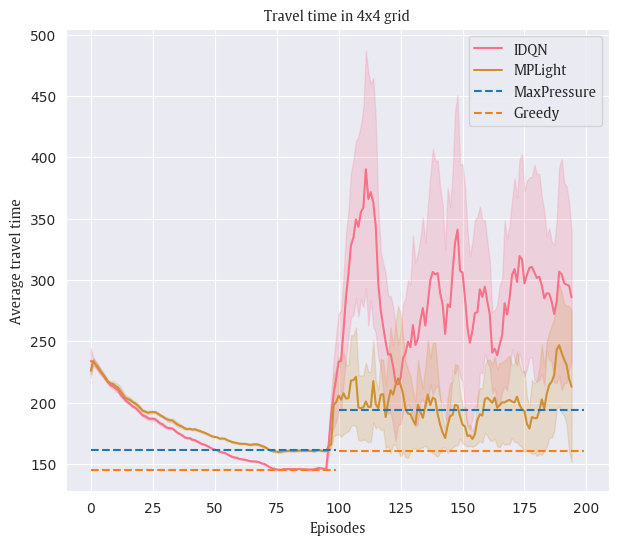}
         \caption{Grid 4x4 network}
         \label{f10a}
     \end{subfigure}
     \hfill
     \begin{subfigure}[b]{0.5\columnwidth}
         \centering
         \includegraphics[width=\columnwidth]{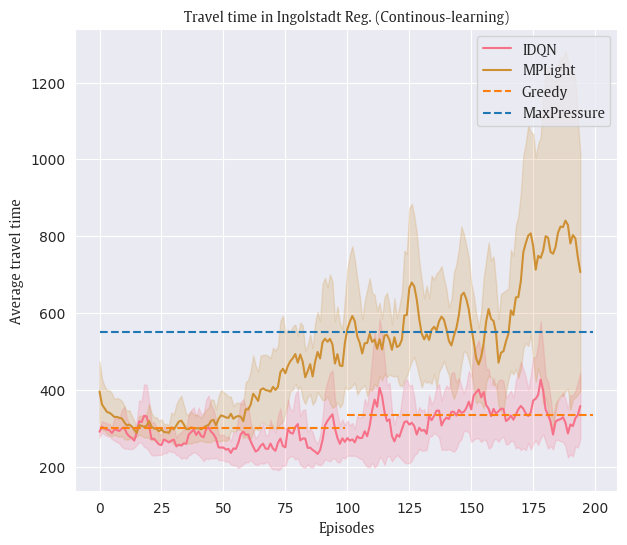}
         \caption{Ingolstadt Region network}
         \label{f10b}
     \end{subfigure}
     \caption{Performance of IDQN and MPLight in the transferability experiment (rolling average over a window of 5 episodes)}
     \label{F10}
\end{figure}

\begin{figure}
     \begin{subfigure}[b]{0.5\columnwidth}
         \centering
         \includegraphics[width=\columnwidth]{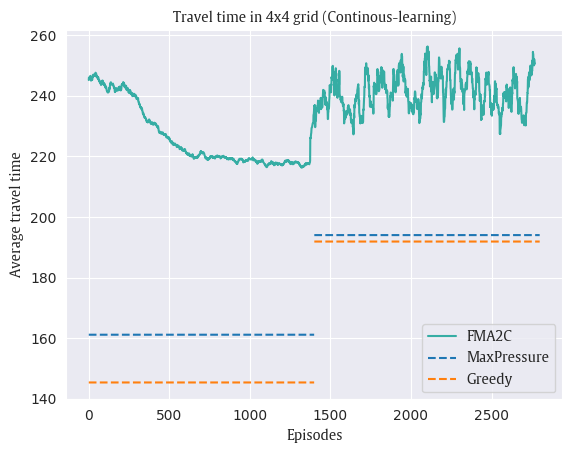}
         \caption{Grid 4x4 network}
         \label{f11a}
     \end{subfigure}
     \hfill
     \begin{subfigure}[b]{0.5\columnwidth}
         \centering
         \includegraphics[width=\columnwidth]{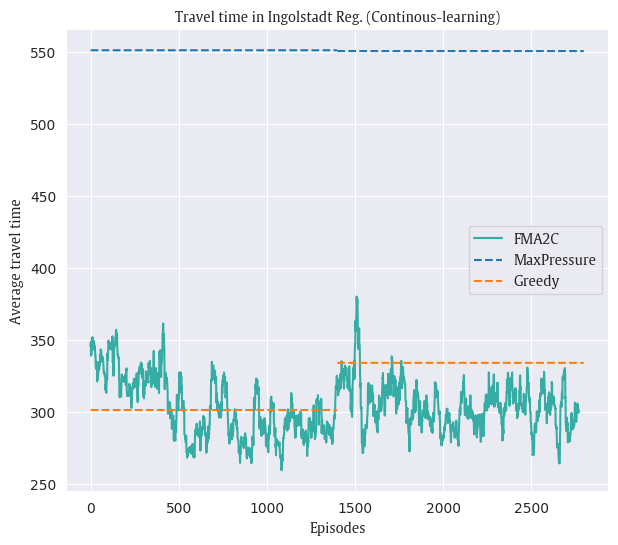}
         \caption{Ingolstadt Region network}
         \label{f11b}
     \end{subfigure}
     \caption{Performance of FMA2C in transferability experiment (rolling average over a window of 30 episodes)}
     \label{F11}
\end{figure}

Figures~\ref{F10} and~\ref{F11} illustrate the average travel time per episode across both the training and transfer phases in Experiment 3.

In the Grid 4×4 network, all three RL-TSC methods exhibit clear convergence during the initial training phase in the base scenario. The convergence levels align with their respective learning paradigms. IDQN, as an independent value-based method, quickly approaches the performance of Greedy through simple exploitation of local patterns, while MPLight, using decentralized coordination via pressure, converges near Max-pressure. FMA2C, which relies on hierarchical coordination and requires more extensive sampling to align manager–worker behaviors, converges the slowest and underperforms, with an average travel time of 217.45 seconds compared to 161.17 for Max-pressure and 145.12 for Greedy.

Upon transfer to the incident scenario, all methods experience a sharp performance drop, but the severity varies by paradigm. IDQN suffers the most substantial increase in travel time—from about 150 seconds to nearly 400 seconds—highlighting the fragility of methods that rely heavily on detailed traffic state representations when exposed to dynamic and previously unseen conditions. In contrast, MPLight and FMA2C show more moderate increases of approximately 60 and 25 seconds, respectively. This smaller spike for MPLight can be attributed to its pressure-based abstraction, which simplifies input features but also reveals overfitting tendencies during continued training. Meanwhile, FMA2C maintains a more stable (although underperforming) trajectory post-transfer, consistent with the higher complexity and sensitivity of hierarchical coordination structures. Notably, IDQN shows signs of adaptation through continued learning, gradually recovering from the initial shock. However, its average performance remains highly volatile. Both MPLight and FMA2C diverge over time in this transfer setting, suggesting limited resilience in handling abrupt environmental changes without additional architectural or mechanisms to capture non-stationary dynamics.

In the Ingolstadt Region network, learning behaviors shift significantly due to the network’s larger scale and structural diversity. Here, MPLight, despite its prior stability in synthetic settings, fails to converge even in the base scenario, and its performance worsens further post-transfer—highlighting the challenge of applying a shared-policy decentralized method in highly heterogeneous networks. IDQN performs more stably during base scenario training, achieving slightly better travel times than Greedy (293.21s vs. 301.70s), but fails to adapt effectively after transfer. Its performance fluctuates heavily under incident conditions, suggesting that while independent value-based learning may generalize somewhat in regular patterns, it struggles to re-optimize when encountering new, disruptive dynamics.

In contrast, FMA2C, representing hierarchical coordination, demonstrates strong adaptability in this larger, more complex setting. It converges gradually in the base scenario and continues improving post-transfer, despite an early performance spike. Its post-transfer performance stabilizes around 303.07 seconds, outperforming Greedy (334.42s) and showing both robustness and continued learning capacity. The architecture’s ability to maintain regional strategy alignment and local responsiveness allows it to outperform other paradigms in this dynamic environment. These results affirm the potential of hierarchical methods to support ongoing adaptation in real-world networks—despite the cost of slower initial convergence and sampling inefficiency.

\section{Conclusion}

In this study, we introduced T-REX, an open-source and reproducible simulation framework designed to train and evaluate RL-TSC methods under incident conditions. Built upon the SUMO platform, T-REX incorporates key behavioral modules—including probabilistic rerouting, speed adaptation, and contextual lane-changing—to realistically model vehicle behavior in disrupted networks. These features enable T-REX to simulate complex, network-level incident dynamics with a high degree of fidelity, supporting the development and benchmarking of robust RL-TSC strategies.

To complement the simulation framework, we proposed a suite of robustness-oriented evaluation metrics that extend beyond conventional traffic efficiency indicators. These metrics provide a more comprehensive understanding of RL-TSC behavior under different real-world deployment paradigms, including pre-training, online learning, and pre-training with continuous learning. Using extensive experiments across both synthetic and real-world networks, we evaluated several state-of-the-art RL-TSC methods in terms of learning efficiency, generalization ability, and adaptability to dynamic traffic conditions, with a particular focus on incident scenarios.

The experimental results reveal that no single RL-TSC paradigm consistently dominates across all networks, scenarios, metrics, and robustness dimensions. Independent value-based methods like IDQN demonstrate fast convergence and strong performance under stable conditions but suffer from overfitting and instability when exposed to dynamic or unseen incident scenarios. This reflects high epistemic uncertainty, as IDQN’s reliance on detailed local states makes it vulnerable to distributional shifts not seen during training. Decentralized pressure-based coordination methods such as MPLight offer better generalization in structured environments due to their abstracted state representation, which helps mitigate aleatoric uncertainty by filtering out stochastic fluctuations. However, they struggle to adapt in heterogeneous networks where local traffic patterns vary widely—revealing limitations in handling epistemic uncertainty. Hierarchical coordination methods like FMA2C exhibit robustness and adaptability, particularly in large-scale or irregular networks, by leveraging structured top-down control. Yet, their performance in mid-sized structured networks is hindered by sensitivity to small regional flow changes, leading to coordination failures and elevated epistemic uncertainty. Moreover, this approach comes at the cost of slower convergence and requires substantially more training interactions with the environment to learn an effective policy. Overall, base-trained policies generalize better than incident-trained ones, suggesting that exposure to disruptions alone is insufficient without addressing the underlying uncertainty. Transferability remains challenging for all methods—highlighting the need for RL-TSC architectures that are explicitly designed to manage both epistemic and aleatoric uncertainties in dynamic real-world traffic environments.

These findings collectively indicate that current state-of-the-art RL-TSC methods are not yet ready for reliable real-world deployment, as they lack robustness to distribution shifts, struggle with out-of-distribution generalization, and fail to consistently handle incident conditions. Bridging this gap will require both architectural innovations and training strategies that explicitly promote generalization and robustness under uncertainty.

Despite its contributions, the T-REX framework has certain limitations. Parameters governing driver awareness and response to incident information are currently inferred or assumed, in the absence of empirical datasets. Future work should aim to integrate more realistic behavioral data to improve the accuracy of rerouting and adaptation modeling. Additionally, T-REX can be extended to support broader robustness assessments, such as sensor failures, fluctuating demand, and adversarial conditions, thereby increasing its utility as a benchmarking platform for real-world RL-TSC deployment.

\section*{Reproducibility Statement}

To support reproducibility, we have made the full source code for the T-REX framework publicly available, including the simulation scripts, traffic network configurations, and the modified SUMO interface used to model incident dynamics. All RL-TSC baselines (IDQN, IPPO, MPLight, and FMA2C) are integrated using openly accessible implementations, with any modifications clearly documented. 

Appendix~\ref{appendix:hyperparams} provides full details of our experimental setup, including the parameter configurations for the \textit{Initializer} and \textit{Deployment} modules, as well as all hyperparameter settings used during training and evaluation. This ensures that our results can be replicated under consistent experimental conditions.

We encourage the research community to use T-REX as a standardized platform for benchmarking RL-TSC methods under real-world disruptions, and to extend its capabilities for future robustness-oriented studies.

\section*{Declarations}






\begin{itemize}
    \item Availability of data and material: The source code and data for the T-REX framework will be made publicly available upon publication. During the review process, all materials will be provided upon request.
    \item Funding: This research was supported by the Independent Research Fund Denmark (Danmarks Frie Forskningsfond) under grant no. 2035-00200B
    \item Acknowledgements: The authors acknowledge the support of the Independent Research Fund Denmark under grant no. 2035-00200B
    \item Competing interests: The authors declare that they have no competing interests.
    \item Authors' contributions: \textbf{Dang Viet Anh Nguyen}: Conceptualization, Methodology, Software, Formal analysis, Writing – original draft. \textbf{Carlos Lima Azevedo}: Supervision, Conceptualization, Methodology, Writing – review and editing. \textbf{Tomer Toledo}: Supervision, Conceptualization, Methodology, Writing – review and editing. \textbf{Filipe Rodrigues}: Supervision, Conceptualization, Methodology, Validation, Writing – review and editing, Funding acquisition.  All authors reviewed and approved the final version of the manuscript.
\end{itemize} 




\begin{appendices}

\section{Rerouting model}\label{secA2}

\subsection{Incident awareness model}

Let \( P_{\text{aware}}(e, t) \) denote the probability that a driver becomes aware of the event while traversing edge \( e \in E \) and exiting at time \( t \). This probability is modeled as a joint function of the complementary probabilities associated with each available source of information \( i \in \{ \text{FTI}, \text{FPI}, \text{OS}, \text{OB} \} \):
\[
P_{\text{aware}}(e, t) = 1 - \prod_{i \in \{\text{FTI}, \text{FPI}, \text{OS}, \text{OB}\}} \left(1 - P_{\text{aware}}^i(e, t)\right).
\]
The probability of awareness through fixed-time information, \( P_{\text{aware}}^{\text{FTI}}(e, t) \), is based on the number of times a driver is exposed to relevant broadcasts while traveling along edge \( e \):
\[
P_{\text{aware}}^{\text{FTI}}(e, t) = \rho^{\text{FTI}}(t + t_e) - \rho^{\text{FTI}}(t),
\]
where \( \rho^{\text{FTI}}(t) \) is the cumulative probability that a driver has become aware of the event via broadcasts up to time \( t \):
\[
\rho^{\text{FTI}}(t) = \mu_{\text{FTI}} \cdot \mu_{\text{ON}} \cdot \sum_{i=1}^{\max(n: t \leq t_n^{\text{FTI}})} (1 - \mu_{\text{ON}})^{i-1}
\]
with \( \mu_{\text{FTI}} \) representing the market penetration of radio listeners, \( \mu_{\text{ON}} \) the probability that a listener notices the event during a broadcast, and \( t_n^{\text{FTI}} \) the time of the \( n \)-th broadcast.

The probability of becoming aware through fixed-place information, \( P_{\text{aware}}^{\text{FPI}}(e, t) \), depends on both spatial and temporal proximity to the information source:
\[
P_{\text{aware}}^{\text{FPI}}(e, t) = \delta_e^{\text{FPI}} \cdot B(t \geq t_{\text{FPI}}) \cdot \frac{1}{1 + \left(\frac{\ell(e, e^d)}{\ell_2}\right)^{\beta^{\text{FPI}}}},
\]
with
\[
\ell(e, e^d) = ||e, e^d|| + \bar{\nu} \cdot ||\text{Mid}(0, t_{\text{start}} - t, t_{\text{end}} - t)||.
\]
In this formulation, \( \delta_e^{\text{FPI}} \) is a binary indicator that equals 1 if the edge \( e \) is equipped with a VMS, and 0 otherwise. The function \( B(\cdot) \) evaluates to 1 if the VMS broadcast has occurred before the driver passes the sign. The function \( \ell(e, e^d) \) denotes the effective distance from the edge to the event location \( e^d \), combining both spatial distance \( ||e, e^d|| \) and a temporal component scaled by the average speed \( \bar{\nu} \). The parameter \( \ell_2 \) defines the maximum effective range of the VMS, and \( \beta^{\text{FPI}} \) controls the rate at which the probability decays with distance. The event duration is represented by \( t_{\text{start}} \) and \( t_{\text{end}} \).

Awareness through online sources is modeled as a cumulative probability function, where the probability of becoming aware while traversing edge \( e \) is:
\[
P_{\text{aware}}^{\text{OS}}(e, t) = \rho^{\text{OS}}(t + t_e) - \rho^{\text{OS}}(t),
\]
and the cumulative probability of awareness at time \( t \) \(\rho^{\text{OS}}(t)\) is defined as:
\[
\rho^{\text{OS}}(t) = \mu_{\text{OS}} \cdot \left[1 - \exp\left(-\frac{(t - t_{\text{OS}})^2}{2\sigma^2}\right)\right].
\]
where \( \mu_{\text{OS}} \) denotes the proportion of drivers using online platforms, \( t_{\text{OS}} \) is the time at which the event was first published online, and \( \sigma \) is a spread-rate parameter governing how quickly the information propagates.

Finally, drivers may become aware of an event through direct observation of abnormal traffic conditions. The probability of observation-based awareness is modeled as:
\[
P_{\text{aware}}^{\text{OB}}(t) = \text{Mid}\left(0, 1, \xi_{\text{OB}} \cdot (t_e - \hat{t}_e)\right) - t_e^{\text{OB}},
\]
where \( t_e \) is the actual travel time on edge \( e \), \( \hat{t}_e \) is the typical (pre-event) travel time, \( t_e^{\text{OB}} \) is the minimum delay required to trigger awareness, and \( \xi_{\text{OB}} \) is a sensitivity parameter that scales the effect of delay on awareness. The function \( \text{Mid}(0, 1, \cdot) \) ensures that the resulting probability is bounded between 0 and 1.

\subsection{Rerouting decision model}

The expected gain, denoted by \( \Delta p^{i}(t) \), captures the change in the utility of available routes by comparing actual edge transition probabilities to typical (pre-event) ones. This is computed based on the cosine similarity between the two probability distributions:
\[
\Delta p^{i}(t) = 1 - \frac{\sum_{e \in E^{+}_{i}} p^{e}(t)\,\hat{p}^{e}(t)}{\sqrt{\sum_{e \in E^{+}_{i}} [p^{e}(t)]^2} \cdot \sqrt{\sum_{e \in E^{+}_{i}} [\hat{p}^{e}(t)]^2}},
\]
where \( p^{e}(t) \) represents the actual transition probability from edge \( i \) to edge \( e \in E^{+}_{i} \) at time \( t \), while \( \hat{p}^{e}(t) \) denotes the corresponding typical (pre-event) transition probability. The set \( E^{+}_{i} \) contains all outgoing edges from edge \( i \) toward the vehicle's destination.

The avoided loss, denoted by \( \Delta w^{i}(t) \), quantifies the improvement in travel cost achieved by rerouting compared to following the typical route. This is calculated as the difference in expected edge potentials (i.e., cost-to-go values) weighted by the transition probabilities:
\[
\Delta w^{i}(t) = \frac{\sum_{e \in E^{+}_{i}} p^{e}(t)\,w^{e}(t) - \sum_{e \in E^{+}_{i}} \hat{p}^{e}(t)\,w^{e}(t)}{\sum_{e \in E^{+}_{i}} \hat{p}^{e}(t)\,w^{e}(t)}.
\]
In this formulation, \( w^{e}(t) \) denotes the minimum expected cost to reach the destination from edge \( e \) at time \( t \).

The total utility of rerouting, \( V_{\text{reroute}} \), is modeled as a linear combination of expected gain and avoided loss:
\[
V_{\text{reroute}} = \beta_{\text{gain}} \cdot \Delta p^{i}(t) + \beta_{\text{loss}} \cdot \Delta w^{i}(t) + \beta_{0},
\]
where \( \beta_{\text{gain}} \) and \( \beta_{\text{loss}} \) are sensitivity parameters that represent the driver's responsiveness to gain and loss, respectively, while $\beta_{0}$ controls the general willingness to reroute.

The resulting rerouting probability at edge \( i \) and time \( t \), denoted by \( \kappa^{i}(t) \), is given by the standard logistic function:
\[
\kappa^{i}(t) = \frac{1}{1 + \exp(-V_{\text{reroute}})}.
\]

\subsection{Routing algorithm}

When a driver decides to reroute, the new path must be computed based on the current network conditions. Although the \texttt{rerouteTravelTime()} function provided by SUMO via TraCI is intended to support dynamic rerouting, it is currently known to be unreliable in certain scenarios \citep{Resumous19:online}. To address this limitation, our framework integrates three alternative online routing algorithms: greedy search, A*, and Dijkstra's algorithm, which provide flexible and accurate route selection under dynamic traffic conditions.

\section{Parameter settings and configuration}\label{secA1}
\label{appendix:hyperparams}
This section outlines the configuration of the \textit{Initializer} and \textit{Deployment} modules in T-REX, along with the hyperparameters used for RL-TSC methods.

\noindent \textbf{Initializer module.} Incident locations and durations are assigned based on network type. For the synthetic \(4 \times 4\) grid, a uniform distribution over all edges is used. In real-world networks, we construct an empirical discrete distribution by simulating one hour of traffic and assigning incident probabilities proportional to observed edge-level traffic volumes. Incident durations follow an exponential distribution \( \text{Exp}(0.029) \), based on real-world incident data from \citep{pereira2013text}.

\noindent \textbf{Deployment module.} Awareness and rerouting behaviors are modeled using multiple information sources:

\begin{itemize}
    \item \textit{Fixed-time information:} Market penetration \( \mu_{FTI} = 0.7 \), detection probability \( \mu_{ON} = 0.5 \), with broadcasts every 5 minutes.
    \item \textit{Fixed-place information (VMS):} Deployed on 40\% of edges, active 10 minutes post-incident \( (t_{FPI} = t_{start} + 10) \), with an effective range of 200 meters and spatial decay \( \beta^{FPI} = 2 \).
    \item \textit{Online sources:} Penetration rate \( \mu_{OS} = 0.8 \), viral delay of 5 minutes, and awareness propagation rate \( \sigma = 10 \) minutes.
    \item \textit{Observation-based awareness:} Triggered when delay exceeds 2 minutes \( (t^{OB}_{e} = 2) \), scaled by sensitivity \( \xi_{OB} = 0.5 \).
\end{itemize}

\noindent \textbf{Rerouting model.} We adopt the utility-based model from \cite{kucharski2019simulation}, with parameters \( \beta_{gain} = 2.5 \), \( \beta_{loss} = 2.5 \), and \( \beta_0 = -5 \). Drivers choose their next edge based on normalized congestion factors, while default route transitions are given probability \( \hat{p}^e(t) = 1 \).

\noindent \textbf{Driver heterogeneity.} We simulate four driver types with varying travel time estimation errors: experienced (40\%, 5\%), novice (30\%, 10\%), distracted (20\%, 20\%), and CAVs (10\%, 1\%). All other parameters follow SUMO defaults.

\noindent \textbf{Lane-changing parameters:} When a vehicle enters its Stopping Sight Distance (SSD) range of a disabled vehicle ahead, its lane-changing behavior is dynamically adjusted in real-time by setting the parameters \texttt{lcStrategic} = 1, \texttt{lcSpeedGain} = 1, \texttt{lcCooperative} = 1, and \texttt{lcKeepRight} = 0. Once the vehicle has passed the disabled vehicle, these parameters are reset to their default values.

\noindent \textbf{RL-TSC methods.} We evaluate IDQN, IPPO, MPLight, and FMA2C using open-source implementations from \cite{ault2021reinforcement}. Default hyperparameters are adopted for IPPO and FMA2C, while IDQN and MPLight undergo additional tuning.

\textit{IDQN:} We conduct a grid search over learning rates \(\{0.01, 0.005, 0.001, 10^{-4}, 10^{-5}, 10^{-6}\}\) and discount factors \(\{0.99, 0.95, 0.90\}\). A learning rate of \(10^{-5}\) performs best for the \(4 \times 4\) grid and Ingolstadt Region; 0.001 is optimal for other networks. Discount factor \( \gamma = 0.99 \) yields the best performance across all environments.

\textit{MPLight:} Optimal learning rates are 0.005 for the grid, 0.01 for Ingolstadt Region, and 0.001 for others, with \( \gamma = 0.99 \).

\noindent \textbf{Phase length.} We test durations of 5 and 10 seconds and observe no significant performance differences. For consistency and training efficiency, we use a fixed 10-second phase length across all experiments \citep{ault2021reinforcement, chen2020toward}.

\section{MDP formulation of RL-TSC methods}\label{A4}

\subsection{IDQN}

\textbf{Observation space $(o^{i})$:} Composed of features extracted from the intersection, including for each lane: 

\begin{itemize}
    \item Current signal assignment (green/yellow/red for each phase),
    \item Number of approaching vehicles,
    \item Number of stopped vehicles,
    \item Accumulated waiting time of stopped vehicles,
    \item Average stop time,
    \item Average speed of approaching vehicles.
\end{itemize}

These are all non-negative and phase-dependent, with a convolutional layer grouping the inputs from lanes that belong to the same road.

\noindent \textbf{Action space $(a^{i})$:} The assignment of the next signal phase (e.g., selecting a non-conflicting phase set) must satisfy two constraints: (1) no conflicting traffic movements, and (2) a yellow phase must be inserted between red and green transitions.

\noindent \textbf{Reward function $(r^{i})$:} Reward is based on the reduction in accumulated traffic delay.

\noindent \textbf{Policy and learning algorithm:} Each agent employs Q-learning (DQN) to approximate its action-value function and selects the phase corresponding to the highest Q-value.

\subsection{IPPO}

Similar to IDQN, but using PPO as the learning algorithm.

\subsection{MPLight}

\textbf{Observation space $(o^{i})$:} Each agent observes:
\begin{itemize}
    \item The current traffic signal phase
    \item The pressure values are defined for each movement through the intersection. Specifically, the movement pressure for $(l, m)$ is given by $x_{(l,m)} = q_{\text{in}} - q_{\text{out}}$, where $q_{\text{in}}$ and $q_{\text{out}}$ are the queue lengths on the entering and exiting lanes, respectively. The phase pressure for a signal phase $s$ is then computed as $p(s) = \sum_{(l,m) \in s} x_{(l,m)}$, representing the total pressure across all permitted movements in that phase.
\end{itemize}

Observations are padded if the intersection has fewer than 12 movements.

\noindent \textbf{Action space $a^{i}$:} Similar to IDQN.

\noindent \textbf{Reward function $r^{i}$:} The negative pressure of intersection $i$ is defined as $r^i = -P_i$, where $P_i$ is the difference between the total entering and exiting queue lengths at intersection $i$.

\noindent \textbf{Policy and learning algorithm:} DQN with FRAP architecture.

\noindent \textbf{Algorithm design:}

\begin{itemize}
    \item Decentralized: Each intersection is controlled by an individual RL agent.
    \item Shared Parameters: All agents share the same model weights to scale efficiently across thousands of intersections.
\end{itemize}

\subsection{FMA2C}

\textbf{Traffic network hierarchy:} Each intersection is controlled by an agent (worker). The network is partitioned into regions, each managed by a manager agent who sets high-level goals for its workers.

\noindent \textbf{MDP for worker agent:}

\noindent \textit{Observation $o^{i}$:} Number of approaching vehicles and cumulative waiting time of the first vehicle at each lane. 

\noindent \textit{Action space $a^{i}$:} Similar to IDQN

\noindent \textit{Reward function $r^{i}$:} Penalizes congestion and waiting time

\noindent \textbf{MDP for manager agent:}

\noindent \textit{Observation $o^{k}$:} Aggregated flows in and out of the region (e.g., directional wave states at region boundaries)

\noindent \textit{Action $a^{k}$:} A high-level goal (desired flow direction) for the region.

\noindent \textit{Reward $r^k$:} Reflects global traffic throughput and liquidity in the region. 

\noindent \textit{Learning algorithm:} Uses Advantage Actor-Critic (A2C) for both manager and worker levels. 

\noindent \textit{Information exchange:} Each worker agent's input includes local and neighbor observations and fingerprints (policy distributions) of neighbors for stability.

\section{Performance of RL-TSC methods on testing environment}\label{A3}

Table~\ref{tab:generalization} presents the detailed performance of RL-TSC methods across different combinations of training and testing scenarios in the generalization experiment. The values reported in the table represent the average travel time.

\begin{sidewaystable}
\caption{Performance of RL-TSC methods on different combinations of training and testing scenarios}
\label{tab:generalization}
\begin{tabular}{lcccccc}
\toprule
\multicolumn{1}{c}{} & \multicolumn{2}{c}{Base - Base}             & Incident - Base      & \multicolumn{2}{c}{Incident - Incident}     & Base - Incident                       \\
\midrule
Grid4x4              & Train (last 10)      & Test (avg. 100)      & Test (avg. 100)      & Train (last 10)      & Test (avg. 100)      & Test (avg. 100)                       \\
\midrule
Max-pressure          & \multicolumn{3}{c}{161.17±1.40}                                   & \multicolumn{3}{c}{194.05±19.68}                                                   \\
IDQN                 & 146.08±1.52          & 152.12±15.73         & 253.69±69.29         & 224.11±70.89         & 239.04±81.84         & 234.57±90.36                          \\
MPLight              & 160.38±1.28          & 159.03±1.55          & 160.60±1.50          & 191.27±23.44         & 179.33±25.28         & 192.02±46.43                          \\
FMA2C                & 217.33±2.39          & 246.90±6.13          & 248.92±6.83          & 239.26±20.49         & 252.30±31.88         & 269.40±27.98                          \\
\midrule
Cologne Cor.         & \multicolumn{1}{l}{} & \multicolumn{1}{l}{} & \multicolumn{1}{l}{} & \multicolumn{1}{l}{} & \multicolumn{1}{l}{} & \multicolumn{1}{l}{}                  \\
\midrule
Max-pressure          & \multicolumn{3}{c}{178.29±93.12}                                  & \multicolumn{3}{c}{266.49±91.25}                                                   \\
IDQN                 & 158.46±220.82        & 130.46±157.98        & 121.93±155.25        & 134.70±41.77         & 260.14±222.51        & 214.86±185.97                         \\
MPLight              & 74.79±4.54           & 124.17±149.12        & 171.63±203.16        & 243.60±133.95        & 247.96±179.14        & 218.07±180.63                         \\
FMA2C                & 129.22±130.39        & 170.89±143.56        & 145.46±92.49         & 183.11±139.71        & 197.52±136.84        & 247.89±177.75                         \\
\midrule
Cologne Reg.         & \multicolumn{1}{l}{} & \multicolumn{1}{l}{} & \multicolumn{1}{l}{} & \multicolumn{1}{l}{} & \multicolumn{1}{l}{} & \multicolumn{1}{l}{}                  \\
\midrule
Max-pressure          & \multicolumn{3}{c}{99.95±14.03}                                   & \multicolumn{3}{c}{133.44±33.95}                                                   \\
IDQN                 & 86.12±0.58           & 85.75±0.48           & 90.08±0.50           & 150.52±101.43        & 146.39±101.89        & 153.26±123.71                         \\
MPLight              & 98.84±2.81           & 92.59±0.67           & 177.88±48.78         & 155.25±84.28         & 223.53±112.81        & 129.32±69.40                          \\
FMA2C                & 98.97±1.83           & 157.86±11.18         & 182.20±19.44         & 139.52±63.85         & 224.02±75.40         & 292.32±84.70                          \\
\midrule
Ingolstadt   Cor.    & \multicolumn{1}{l}{} & \multicolumn{1}{l}{} & \multicolumn{1}{l}{} & \multicolumn{1}{l}{} & \multicolumn{1}{l}{} & \multicolumn{1}{l}{}                  \\
\midrule
Max-pressure          & \multicolumn{3}{c}{80.62±2.30}                                    & \multicolumn{3}{c}{137.14±43.41}                                                   \\
IDQN                 & 72.64±1.02           & 73.41±1.95           & 81.16±3.63           & 163.40±95.14         & 173.61±128.51        & 184.58±167.66                         \\
MPLight              & 101.01±62.35         & 82.81±16.34          & 85.29±14.39          & 149.18±48.54         & 174.06±124.25        & 153.24±126.79                         \\
FMA2C                & 90.65±9.76           & 140.90±18.09         & 153.47±19.70         & 151.38±87.34         & 208.52±93.25         & 189.37±94.11                          \\
\midrule
Ingolstadt   Reg.    & \multicolumn{1}{l}{} & \multicolumn{1}{l}{} & \multicolumn{1}{l}{} & \multicolumn{1}{l}{} & \multicolumn{1}{l}{} & \multicolumn{1}{l}{}                  \\
\midrule
Max-pressure          & \multicolumn{3}{c}{581.72±63.83}                                  & \multicolumn{3}{c}{596.90±57.83}                                                   \\
IDQN                 & 72.64±1.02           & 73.41±1.95           & 273.01±84.10         & 302.722±81.986       & 453.238±184.311      & 284.89±97.19                          \\
MPLight              & 101.01±62.35         & 82.81±16.34          & 563.39±123.11        & 343.379±46.331       & 555.62±156.836       & 515.20±116.72                         \\
FMA2C                & 90.65±9.76           & 140.90±18.09         & 356.02±79.09         & 297.13±71.23         & 334.26±90.61         & 366.08±63.04             \\
\bottomrule
\end{tabular}
\end{sidewaystable}




\end{appendices}


\bibliography{sn-bibliography}

\end{document}